\documentclass[11pt]{article}

\PassOptionsToPackage{table}{xcolor}
\usepackage[preprint]{acl}

\usepackage{times}
\usepackage{latexsym}
\usepackage[T1]{fontenc}
\usepackage[utf8]{inputenc}
\usepackage{microtype}
\usepackage{inconsolata}
\usepackage{graphicx}
\usepackage{multicol}
\usepackage{url}

\usepackage{amsmath,amsfonts,bm}

\def\eqref#1{equation~\ref{#1}}

\def\1{\bm{1}}

\DeclareMathAlphabet{\mathsfit}{\encodingdefault}{\sfdefault}{m}{sl}
\SetMathAlphabet{\mathsfit}{bold}{\encodingdefault}{\sfdefault}{bx}{n}

\usepackage{amsmath}
\usepackage{amssymb}

\usepackage{tikz}
\usepackage{pifont}
\usetikzlibrary{arrows.meta,positioning}

\usepackage{booktabs}
\usepackage{multirow}
\usepackage{tabularx}
\usepackage{array}

\usepackage{subcaption}

\usepackage{algorithm}
\usepackage{algorithmic}
\usepackage{placeins}
\usepackage{xurl}
\usepackage[most]{tcolorbox}
\usepackage{tcolorbox}

\definecolor{metateamblue}{RGB}{235,243,252}

\newtcolorbox{promptbox}[1]{
    enhanced,
    breakable,
    width=\linewidth,
    colback=gray!8,
    colframe=black!70,
    boxrule=0.8pt,
    arc=2mm,
    left=3mm,
    right=3mm,
    top=2.5mm,
    bottom=2.5mm,
    before skip=6pt,
    after skip=6pt,
    fontupper=\small,
    before upper={
        \textbf{#1}\par
        \vspace{1mm}
    }
}



\title{
EvoTeam: Multi-Agent Collaboration via Evolving Teams
}

\author{
  \textbf{Qinuo Cao}$^{1}$\thanks{Equal contribution.},
  \textbf{Xin He}$^{1}$\footnotemark[1],
  \textbf{Dianxuan Fu}$^{1}$,
  \textbf{Lizhi Zhang}$^{1}$,
  \textbf{Zhaoqing Xing}$^{1}$,
  \textbf{Jiapu Wang}$^{2}$,\\
  \textbf{Xin Wang}$^{1}$\thanks{Corresponding authors.},
  \textbf{Na Zou}$^{3}$,
  \textbf{Yi Chang}$^{1}$,
  \textbf{Qinggang Zhang}$^{1}$\footnotemark[2]
  \\[2pt]
  $^{1}$Jilin University
  $^{2}$Nanjing University of Science and Technology
  $^{3}$Shanghai AI Lab
  \\[2pt]
    \texttt{\{caoqn26,hex23\}@mails.jlu.edu.cn};\;
  \texttt{\{qinggangzhang,xinwang\}@jlu.edu.cn}
}

\begin{document}

\maketitle

\begin{abstract}
LLM-based multi-agent systems (MAS) rely on collaboration among specialized agents to solve complex tasks. Existing methods improve MAS by refining the prompts, skills, and execution strategies of individual agents based on task feedback. However, improving individual agents does not necessarily lead to better team collaboration. Recent studies have therefore extended evolution to the collaboration level, using collective execution experience to adjust team organization, and inter-agent interactions. Nevertheless, evolving collaboration across heterogeneous tasks remains challenging, as existing approaches often constrain team reorganization within a shared structure and reuse interaction rules across incompatible team contexts. 
In this paper, we propose \textsc{EvoTeam}, a collaboration-centric evolution framework that jointly evolves team organization, interaction rules, and individual agent skills through execution experience. It maintains a library of reusable teams, selecting or creating agent teams and guiding collaboration through team-specific interaction rules. Collaboration experience accumulates within compatible team contexts, while individual skills remain reusable across teams. Experiments across four benchmarks demonstrate its effectiveness. Our code is available at \url{https://github.com/DEEP-JLU/EvoTeam}.
\end{abstract}

\section{Introduction}
\label{sec:introduction}
\vspace{-3mm}
Large language model (LLM)-based multi-agent systems (MAS) organize specialized agents into collaborative teams to effectively combine their complementary capabilities and accomplish complex tasks that require diverse expertise, sustained reasoning, and extensive tool interactions~\citep{feng2026graph,chen2023agentversefacilitatingmultiagentcollaboration}.
Such collaboration has enabled complex applications like software engineering and multi-step decision-making~\citep{wu2023autogenenablingnextgenllm,qian-etal-2024-chatdev}. As these applications expand, enabling MAS to continuously improve through accumulated execution experience has become an important research direction~\citep{hao2026recreatereasoningcreatingdomain,lin2026skillmasevolvingmetaskillautomatic}.

To enable such continual improvement, existing research has explored experience-driven evolution in multi-agent systems. A prominent direction focuses on improving the capabilities of individual agents through accumulated execution experience. By learning from previous task trajectories and feedback, agents can refine their prompts, acquire reusable skills, and improve their reasoning and tool-use strategies~\citep{hao2026recreatereasoningcreatingdomain,zheng2025skillweaverwebagentsselfimprove}. These improvements can be retained across tasks, allowing individual agents to progressively strengthen their task-solving capabilities~\citep{li2025crosstaskexperientiallearningllmbased}. Such experience-driven approaches provide an important foundation for MAS evolution by enabling agents to learn from previous executions rather than repeatedly solving tasks from scratch.

However, evolving better individual agents does not necessarily produce better collaborative teams. Multi-agent collaboration depends on how complementary capabilities are organized and coordinated toward a shared objective~\citep{feng2026graph,zhou2026multiagentdesignoptimizingagents}. Even when agents possess necessary expertise, inappropriate responsibility assignments may prevent their capabilities from being effectively utilized, while incomplete information handoffs or insufficient verification may undermine correct individual contributions~\citep{cemri2025multiagentllmsystemsfail}. Effective MAS evolution must extend beyond improving what individual agents can do to improving how they organize, communicate, and collaborate~\cite{ruan2026aorchestraautomatingsubagentcreation,feng2026graph}.

Recent studies have begun to extend experience-driven evolution from individual agents to multi-agent collaboration. Rather than refining agent capabilities in isolation, these approaches leverage collective execution experience to adapt team organization, adjust agent responsibilities, and improve inter-agent interactions~\citep{hao2026evolveteamcollaborativeselfevolution,pan2026skillmasskillcoevolutionllmbased,fei2026emasstabilizingmultiagentevolution}. However, different tasks may require distinct combinations of agents, divisions of responsibility, and interaction rule~\citep{yue2025masrouter,pmlr-v267-zhang25bi,kim2026sciencescalingagentsystems}. Collaboration patterns learned from previous executions may not remain appropriate as task requirements change. This raises a critical question: \textit{how can MAS continuously evolve collaboration while adapting to heterogeneous tasks and preserving useful experience?}

\begin{figure*}[t]
    \centering
    \begin{subfigure}[t]{0.475\linewidth}
        \centering
        \includegraphics[width=\linewidth]{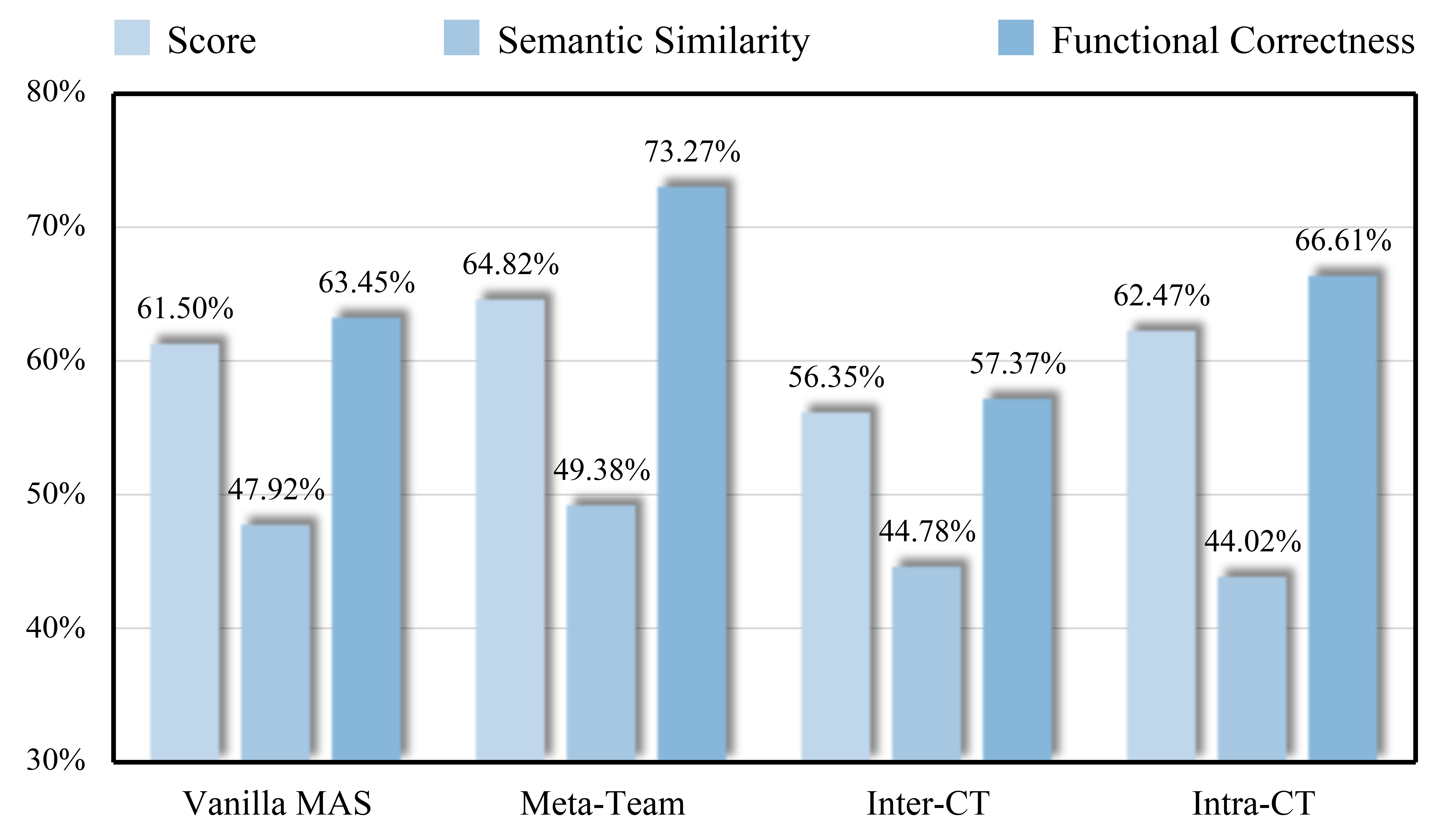}
        \caption{Target-task performance of Vanilla MAS, target-only \textsc{Meta-Team}, and two CT settings.
        }
        \label{fig:preliminary_study}
    \end{subfigure}\hfill%
    \begin{subfigure}[t]{0.475\linewidth}
        \centering
        \includegraphics[width=\linewidth]{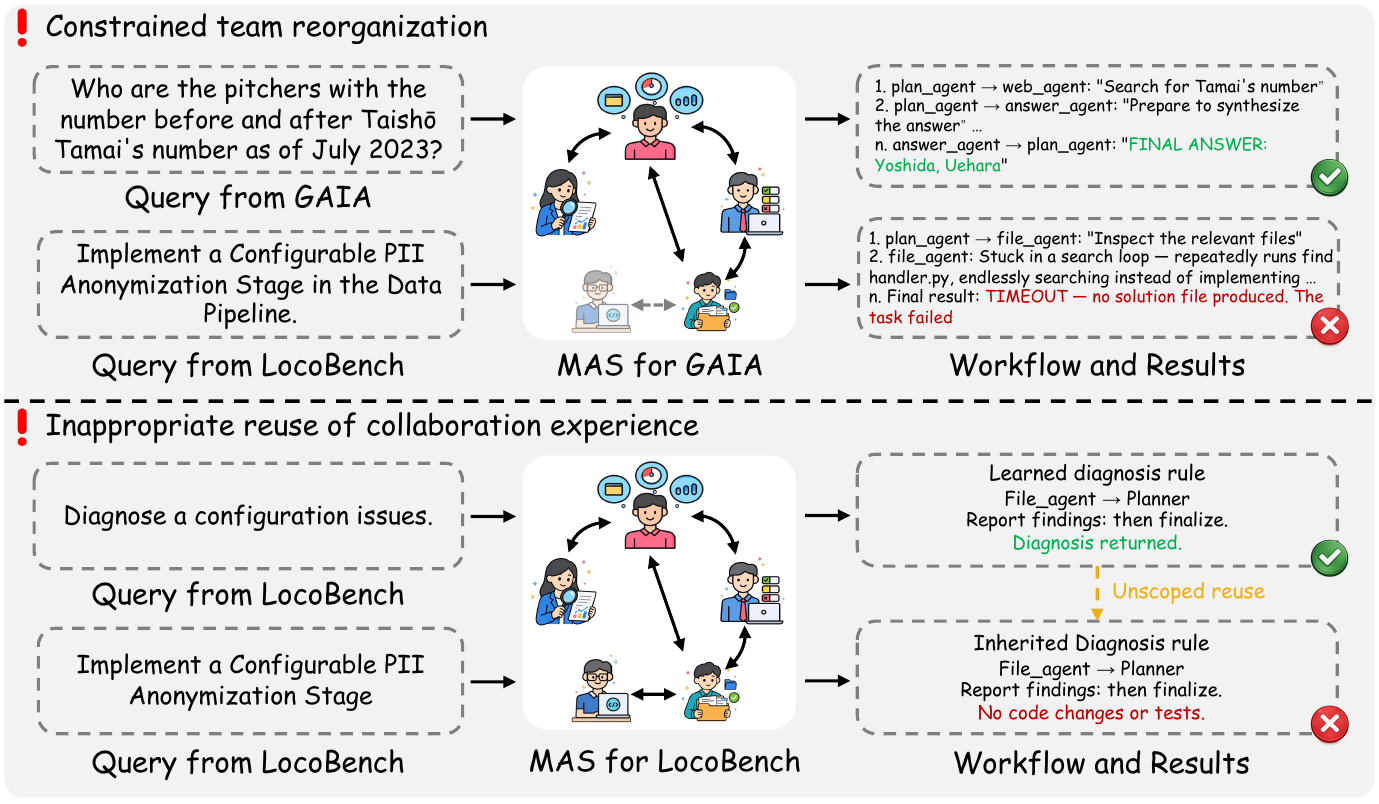}
        \caption{Representative failure modes of ineffective collaboration experience reuse across heterogeneous tasks.
        }
        \label{fig:error-analysis}
    \end{subfigure}
    \vspace{-2mm}    
    \caption{Preliminary study: experience reuse in MAS.
(a) Cross-task training underperforms target-only training on target tasks.
(b) Trajectories show how inflexible templates and unclear rule scopes hinder reuse.}
    \label{fig:motivation}
    \vspace{-6mm}
\end{figure*}

Despite recent advances, existing approaches face two critical limitations in collaboration evolution. \textbf{(i) Constrained team reorganization.} Existing approaches often adjust agent membership and responsibilities within a shared team, limiting their ability to maintain distinct collaboration patterns for tasks with different requirements. \textbf{(ii) Inappropriate reuse of collaboration experience.} Interaction rules learned from previous tasks are often tied to specific agent responsibilities and information dependencies. Reusing or updating these rules across incompatible collaboration contexts may introduce unsuitable handoffs or stopping conditions.
Our preliminary study in Fig~\ref{fig:motivation} further illustrates these difficulties: continued evolution across heterogeneous tasks can reduce task performance relative to target-only evolution, while representative execution trajectories reveal mismatches in team organization and interaction rule applicability.

\begin{figure}[t]
    \centering
    \includegraphics[width=\columnwidth]{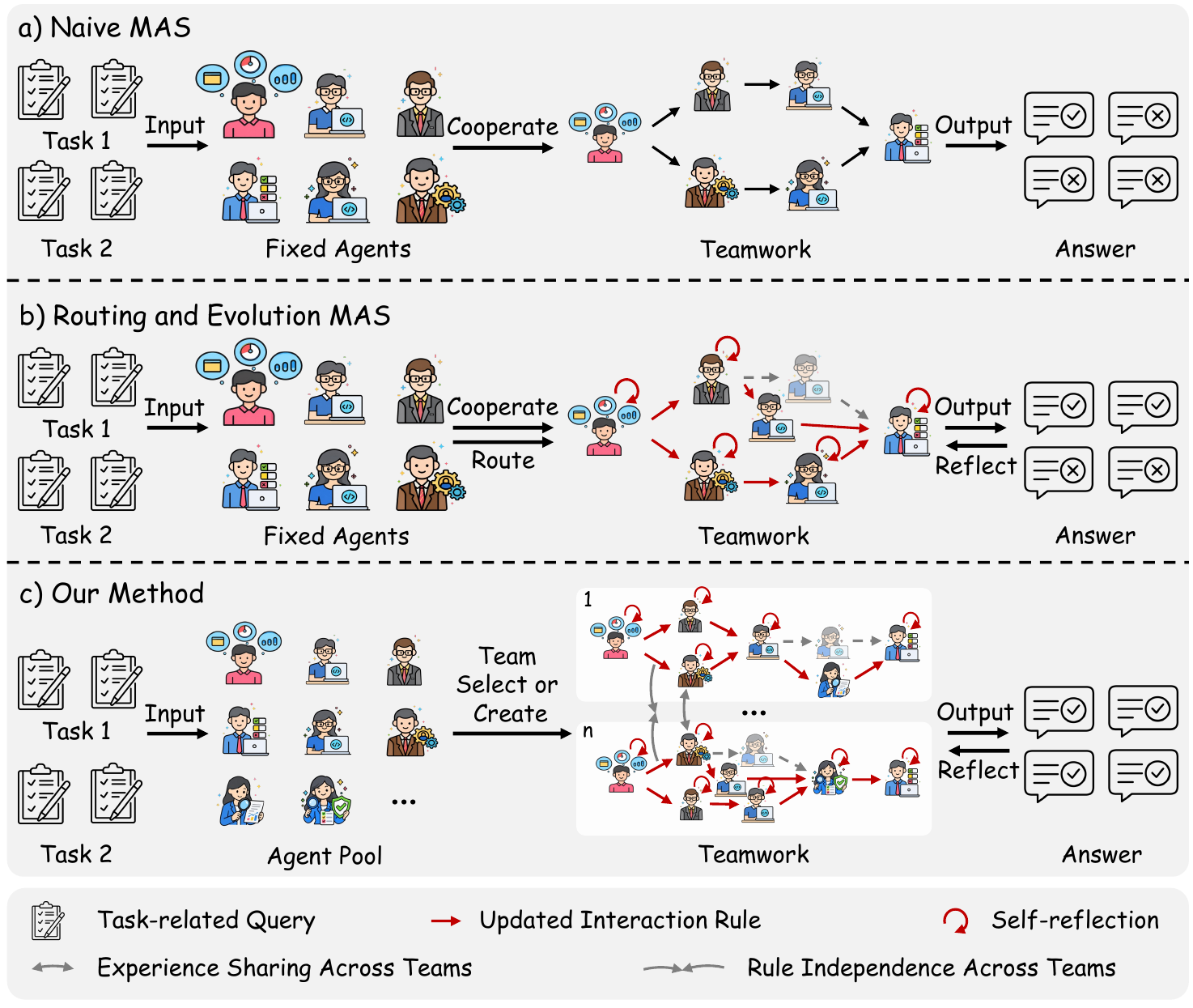}
    \vspace{-8mm}
    \caption{Three paradigms of MAS.}
    \label{fig:mas-paradigms}
    \vspace{-8mm}
\end{figure}
In this paper, we propose \textsc{EvoTeam}, a collaboration-centric evolution framework that jointly evolves team organization, interaction rules, and individual agent skills through execution experience. \textsc{EvoTeam} organizes collaboration experience around task-appropriate teams, allowing compatible tasks to reuse and refine interaction rules while preserving individual agent skills across teams. Specifically, \textsc{EvoTeam} consists of three key stages. \textbf{(i) Adaptive Agent Team Composition} maintains a library of agent teams, selecting a compatible team or creating a new one according to task requirements.  \textbf{(ii) Team-Scoped Collaboration} coordinates task execution through task-specific responsibility assignments and applicable team-specific interaction rules, allowing tasks with compatible collaboration requirements to reuse accumulated coordination experience while keeping rule updates independent across teams. \textbf{(iii) Multi-Level Experience Evolution} leverages execution traces and task feedback to refine team organization, update team-specific interaction rules, and improve individual agent skills. While collaboration experience evolves within corresponding teams, individual skills remain reusable across team contexts. 
Our contributions are summarized as follows:
\begin{itemize}
    \item 
    We identify interconnected challenges on collaboration-centric multi-agent evolution and propose \textsc{EvoTeam}, a framework that maintains and evolves reusable teams and their interaction rules while preserving individual agent skills across team contexts.
    \item  \textsc{EvoTeam} consists three key components that enables team organization and coordination rules to improve jointly through accumulated execution experience.
    \item We conduct experiments across four benchmarks and six task categories, demonstrating the effectiveness of \textsc{EvoTeam} under individual-benchmark and mixed-benchmark evolution. Ablation studies, cross-backbone evaluations, and case analyses examine its collaboration mechanisms and experience reuse.
\end{itemize}

\vspace{-3mm}
\section{Preliminary Study}\label{sec:preliminary}
\vspace{-2mm}
In this section, we investigate whether experience accumulated on previous tasks benefits continued evolution of MAS across heterogeneous tasks. We compare performance across evolution settings and examine how team organization and interaction-rule applicability may limit experience reuse.

\subsection{Cross-Task Performance Degradation}
In our preliminary experiments, \textsc{Meta-Team} exhibits lower target-task performance after cross-task evolution than after target-only evolution.
To examine this degradation, we evaluate two cross-task (CT) evolution settings.
In both settings, the system inherits an evolved MAS and continues its evolution on the target training tasks.
We use target-only \textsc{Meta-Team} as the primary reference and Vanilla MAS as a non-evolutionary reference.
Inter-CT starts from another benchmark with an unsuitable team template, whereas Intra-CT starts from another task category within the same benchmark with a compatible team template.
As shown in Figure~\ref{fig:preliminary_study}, both settings underperform target-only \textsc{Meta-Team} on all three metrics.
These results suggest that prior experience does not necessarily benefit continued evolution, even when a compatible team template is retained.

\subsection{Analysis of Potential Failure Modes}
\label{sec:preliminary_analysis}
Figure~\ref{fig:error-analysis} illustrates two potential sources of the observed degradation.
(i) Constrained team reorganization.
A source-oriented team template may not support the division of labor required by the target tasks.
The GAIA-oriented team coordinates web retrieval and answer synthesis but fails to establish an effective implementation and testing process for the LocoBench PII-anonymization task.
(ii) Inappropriate reuse of interaction rules.
Even with a suitable team template, inherited interaction rules may prescribe inappropriate handoffs or stopping conditions.
In the lower panel, a diagnosis-oriented rule directs the file agent to report its findings to the planner and finalize the answer.
If this rule persists during adaptation to an implementation task, the system may terminate before coding and testing.


\subsection{Discussion}
Our preliminary study suggests that retaining prior experience alone is insufficient for effective MAS evolution across heterogeneous tasks.
Although accumulated agent skills may remain useful across team contexts, reusing the original team template and interaction rules without accounting for new task requirements can constrain subsequent evolution.
\textbf{(i) Team adaptation should maintain distinct collaboration structures.}
Selecting or creating task-appropriate teams enables reorganization to meet changing collaboration requirements.
\textbf{(ii) Interaction rules should be shared and refined only among tasks with compatible collaboration requirements.}
A clear reuse scope helps prevent rule updates for one task type from disrupting the coordination required by others.
Guided by these principles, we develop \textsc{EvoTeam} to support the accumulation and selective reuse of collaboration experience across heterogeneous tasks.



\vspace{-3mm}
\section{The Framework of \textsc{EvoTeam}}
\label{sec:method}
\vspace{-2mm}

\begin{figure*}[!t]
  \centering
  \includegraphics[width=\textwidth]
  {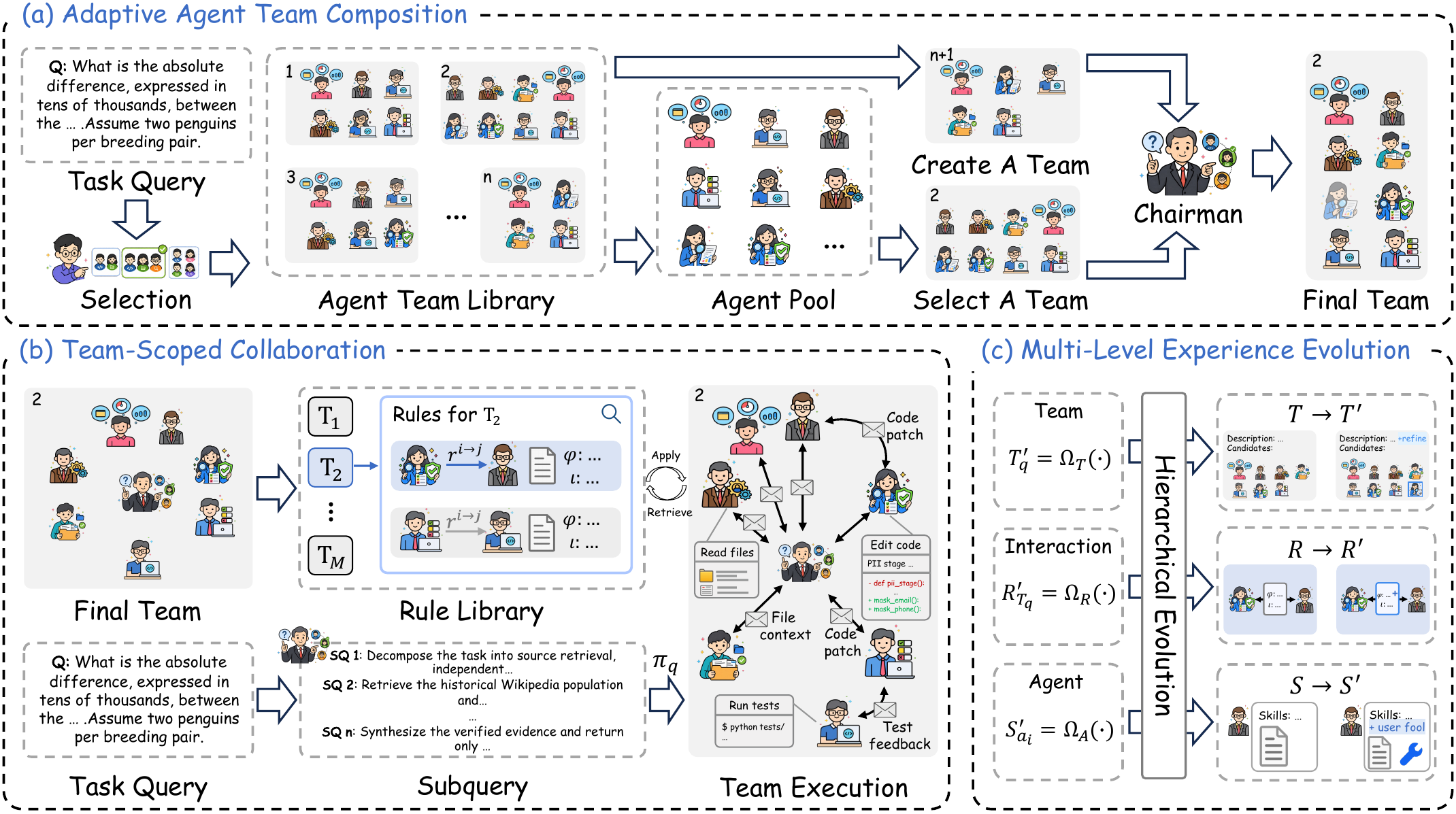}
  \vspace{-6mm}
  \caption{\textbf{The overall framework of \textsc{EvoTeam}.}
\textbf{(a) Adaptive Agent Team Composition.} The system selects or creates a team, and the Chairman recruits agents from its candidate roster. \textbf{(b) Team-Scoped Collaboration.} Agents execute assigned subtasks using accumulated skills and applicable team-specific rules. \textbf{(c) Multi-Level Experience Evolution.} Reflectors refine teams, team-specific rules, and agent skills using execution traces.}\label{fig:framework}
\vspace{-6mm}
\end{figure*}


Our preliminary study indicates that experience reuse across heterogeneous tasks requires flexible team formation and selective reuse of interaction rules. Motivated by these requirements, we develop \textsc{EvoTeam}, a framework that selects or creates task-appropriate agent teams to reorganize agents and evolves interaction rules within each team. Figure~\ref{fig:framework} summarizes its three stages per task. \textbf{(i) Adaptive Agent Team Composition:} \textsc{EvoTeam} selects or creates a task-appropriate agent team and recruits agents from its candidate roster. \textbf{(ii) Team-Scoped Collaboration:} Agents use accumulated skills to complete subtasks and consult the selected agent team's interaction rules. \textbf{(iii) Multi-Level Experience Evolution:} Execution evidence refines agent teams, updates team-specific interaction rules, and evolves agent skills for reuse across agent teams.

\subsection{Adaptive Agent Team Composition}
\label{sec:method_template}
\label{sec:template-selection}

A fixed team can constrain agent reorganization when task requirements vary.
To decouple agent capabilities from team organization, \textsc{EvoTeam} maintains a global agent pool \(\mathcal{A}=\{a_1,\dots,a_N\}\) and a library of reusable agent teams \(\mathcal{T}=\{T_1,\dots,T_M\}\).
An agent team \(T=\langle d_T,\mathcal{C}_T\rangle\) represents a reusable team organization, where \(d_T\) describes the collaboration requirements it supports and \(\mathcal{C}_T\subseteq\mathcal{A}\) specifies its candidate agent roster.
For each task, an execution team is recruited from this roster according to the task's functional requirements.

\paragraph{Team Creation.}
When \(\mathcal{T}=\emptyset\) or no existing agent team satisfies the collaboration requirements of the current query \(q\), an LLM-based \textit{Composer} constructs a new provisional agent team from the available agent pool: \(\tilde{T}_q=\Phi_C(q,\mathcal{A})\).
The provisional team is used for the current task, and its admission to \(\mathcal{T}\) is determined after task completion.

\paragraph{Team Selection.}
For each agent team \(T\in\mathcal{T}\), an LLM-based \textit{Team Selector} \(\Phi_S\) checks whether \(d_T\) covers the collaboration requirements of the current query \(q\) and whether \(\mathcal{C}_T\) contains agents with the required functional roles.
The agent team for query \(q\) is selected or created as follows:
 \begin{equation}
     T_q =
    \begin{cases}
        \Phi_S(q, \mathcal{T}), & \text{if } \Phi_S(q, \mathcal{T}) \neq \emptyset, \\[2pt]
        \tilde{T}_q, & \text{otherwise},
     \end{cases}
 \end{equation}
where \(\Phi_S(q,\mathcal{T})=\emptyset\) indicates that no compatible agent team is available in the library. 
Tasks can consequently reuse the same agent team when they require an analogous division of work among agents, even if their subject matter differs.

\paragraph{Agent Recruitment.}
The selected agent team \(T_q\) provides a candidate roster \(\mathcal{C}_{T_q}\) suited to the task's collaboration requirements.
Within this roster, an LLM-based \textit{Chairman} \(\Phi_{\text{Chair}}\) recruits the agents for the query \(q\) by matching its functional requirements to the candidate agents' specialized roles:
\begin{equation}
E_q=\Phi_{\text{Chair}}(q,T_q),
\qquad E_q\subseteq\mathcal{C}_{T_q},
\end{equation}
where \(E_q\) denotes the task-specific execution team. Tasks that reuse the same agent team can recruit different subsets of its candidate roster. This task-specific recruitment is intended to reduce unnecessary participation, interaction, and token use.

\subsection{Team-Scoped Collaboration}
\label{sec:task-execution}

Interaction rules learned under a different agent team may not align with agents' current responsibilities and coordination needs, providing ineffective guidance or even hindering task execution.
\textsc{EvoTeam} maintains team-specific interaction rule sets to narrow the scope of reuse.
The selected agent team \(T_q\) determines the available interaction rule set \(\mathcal{R}_{T_q}\), while recruited agents assess the applicability of individual rules to their current coordination needs.
Each recruited agent \(a_i\in E_q\) executes its assigned subtasks using its accumulated skill set \(S_{a_i}\) and consults \(\mathcal{R}_{T_q}\) in its context to guide coordination with other agents.

\paragraph{Intra-Team Planning.}
Given the query \(q\), the selected agent team \(T_q\), and the recruited execution team \(E_q\), the Chairman decomposes the task into a set of subtasks \(U_q=\{u_1,\dots,u_m\}\) and establishes a corresponding subtask assignment plan \(\pi_q\):
\begin{equation}
    (U_q, \pi_q) = \Phi_{\text{Chair}}(q, T_q, E_q),
\end{equation}
where \(\pi_q: U_q \to E_q\) assigns each subtask \(u \in U_q\) to an agent in \(E_q\). 
The assignment relies on subtask requirements and agent profiles, which summarize agents' areas of expertise and reliability. 
The plan specifies subtask responsibilities for this task without predetermining interaction partners or interaction rules used during collaboration.

\paragraph{Collaborative Team Execution.}
Guided by \(\pi_q\), agents execute their assigned subtasks using their accumulated skills and coordinate with peers when additional information or intermediate results are needed. 
The interaction rule set \(\mathcal{R}_{T_q}\) associated with the selected agent team is injected into their contexts as soft coordination guidance. 
During coordination, agents select a suitable rule based on both its description and the current coordination requirements to guide interaction and information handoffs between participating agents.

\paragraph{Team Execution Traces.}
The Chairman monitors subtask progress during task execution and adjusts \(\pi_q\) by reassigning subtasks within \(E_q\) when the current allocation of responsibilities blocks progress. 
The system records individual agent traces and an inter-agent handoff trace, which jointly form the composite execution trace:
\begin{equation}
\tau_c=
\left\langle
\{\tau_{a_i}\}_{a_i\in E_q},
\tau_t
\right\rangle,
\end{equation}
where \(\tau_{a_i}\) records \(a_i\)'s reasoning steps, tool outputs, and intermediate artifacts. 
The handoff trace \(\tau_t\) records interaction events, transferred payloads, and interaction rules used for coordination.

\subsection{Multi-Level Experience Evolution}
\label{sec:method_evolution}
\textsc{EvoTeam} distills execution experience into reusable knowledge at the team, interaction, and agent levels.
The corresponding reflectors use \(\tau_c\), \(\tau_t\), and \(\tau_{a_i}\), respectively.
The updates concern agent team descriptions or candidate rosters, interaction rules, and individual agent skills, respectively.
Interaction rules remain team-specific, whereas agent skills remain available for reuse across agent teams.

\paragraph{Team-level evolution.}
The Team Reflector \(\Omega_T\) uses the query \(q\) and the composite trace \(\tau_c\) to assess whether the selected agent team's candidate roster supports the required division of work:
\begin{equation}
T_q'=\Omega_T(T_q,\tau_c,q),
\end{equation}
where \(T_q'\) is the resulting team.
When the assessment identifies a need to change the candidate roster, the reflector revises it and updates the agent team's description accordingly.
The revised description specifies the collaboration requirements supported by the updated roster.
For a newly created agent team, admission to \(\mathcal{T}\) requires both a candidate roster that supports the required division of work and a reusable collaboration structure.

\paragraph{Interaction-level evolution.} 
The Rule Reflector \(\Omega_R\) uses query \(q\) and handoff trace
\(\tau_t\) to update the rule set associated with the selected agent team \(T_q\):
\begin{equation}
    \mathcal{R}_{T_q}' = \Omega_R(\mathcal{R}_{T_q}, \tau_t, q),
\end{equation}
where \(\mathcal{R}_{T_q}'\) is the revised rule set. 
The reflector examines interaction events between participating agents, transferred payloads, and the use of interaction rules to identify missing information or mismatches between handoff guidance and the agents' responsibilities. 
These observations guide the creation of new interaction rules, the revision of existing rules, or the removal of rules that no longer fit the agent team's collaboration requirements.

\begin{table*}[t]
\caption{Main results across diverse agent benchmarks under both individual-benchmark and mixed-benchmark evolution. Subscripts show absolute score differences from SA. The best scores within each block are in \textbf{bold}.}
\vspace{-2mm}
\centering
\begingroup
\small
\setlength{\tabcolsep}{3pt}
\renewcommand{\arraystretch}{1.22}
\definecolor{mainup}{RGB}{200,40,40}
\definecolor{maindown}{RGB}{36,95,165}
\definecolor{metateamblue}{RGB}{235,243,252}
\newcommand{\maincell}[2]{%
  \makebox[2.65em][r]{#1}%
  \hspace{0.12em}%
  \makebox[2.7em][l]{\textsubscript{\scriptsize #2}}%
}
\resizebox{\textwidth}{!}{%
\begin{tabular}{@{}l*{7}{c}@{}}
\toprule
\multirow{2}{*}{\textbf{Method}}
& \textbf{GAIA}
& \multicolumn{2}{c}{\textbf{LoCoBench}}
& \textbf{LOCA}
& \multicolumn{2}{c}{\textbf{BeyondSWE}}
& \multirow{2}{*}{\textbf{Avg.}} \\
\cmidrule(lr){2-2}\cmidrule(lr){3-4}\cmidrule(lr){5-5}\cmidrule(lr){6-7}
& \textbf{Val} & \textbf{Feat.} & \textbf{Refact.}
& \textbf{Val} & \textbf{DepMig.} & \textbf{CrossR.} & \\
\midrule
\rowcolor{gray!12}
\multicolumn{8}{@{}l@{}}{\textbf{Individual-Benchmark Evolution}} \\

SA
& \maincell{64.00}{}
& \maincell{55.99}{}
& \maincell{53.08}{}
& \maincell{37.50}{}
& \maincell{34.81}{}
& \maincell{37.22}{}
& \maincell{47.10}{} \\

MAS
& \maincell{72.00}{\textcolor{mainup}{$\uparrow\!8.00$}}
& \maincell{61.71}{\textcolor{mainup}{$\uparrow\!5.72$}}
& \maincell{61.49}{\textcolor{mainup}{$\uparrow\!8.41$}}
& \maincell{49.17}{\textcolor{mainup}{$\uparrow\!11.67$}}
& \maincell{29.75}{\textcolor{maindown}{$\downarrow\!5.06$}}
& \maincell{38.33}{\textcolor{mainup}{$\uparrow\!1.11$}}
& \maincell{52.08}{\textcolor{mainup}{$\uparrow\!4.98$}} \\

OWL
& \maincell{48.00}{\textcolor{maindown}{$\downarrow\!16.00$}}
& \maincell{51.80}{\textcolor{maindown}{$\downarrow\!4.19$}}
& \maincell{50.64}{\textcolor{maindown}{$\downarrow\!2.44$}}
& \maincell{39.17}{\textcolor{mainup}{$\uparrow\!1.67$}}
& \maincell{33.54}{\textcolor{maindown}{$\downarrow\!1.27$}}
& \maincell{39.44}{\textcolor{mainup}{$\uparrow\!2.22$}}
& \maincell{43.77}{\textcolor{maindown}{$\downarrow\!3.33$}} \\

AgentSquare
& \maincell{49.33}{\textcolor{maindown}{$\downarrow\!14.67$}}
& \maincell{62.45}{\textcolor{mainup}{$\uparrow\!6.46$}}
& \maincell{60.55}{\textcolor{mainup}{$\uparrow\!7.47$}}
& \maincell{43.33}{\textcolor{mainup}{$\uparrow\!5.83$}}
& \maincell{34.81}{\raisebox{0.6ex}{\hspace{0.3em}\textemdash}}
& \maincell{35.56}{\textcolor{maindown}{$\downarrow\!1.66$}}
& \maincell{47.67}{\textcolor{mainup}{$\uparrow\!0.57$}} \\

Skill-MAS
& \maincell{48.67}{\textcolor{maindown}{$\downarrow\!15.33$}}
& \maincell{64.38}{\textcolor{mainup}{$\uparrow\!8.39$}}
& \maincell{62.57}{\textcolor{mainup}{$\uparrow\!9.49$}}
& \maincell{51.25}{\textcolor{mainup}{$\uparrow\!13.75$}}
& \maincell{34.81}{\raisebox{0.6ex}{\hspace{0.3em}\textemdash}}
& \maincell{33.33}{\textcolor{maindown}{$\downarrow\!3.89$}}
& \maincell{49.17}{\textcolor{mainup}{$\uparrow\!2.07$}} \\

ReCreate
& \maincell{56.00}{\textcolor{maindown}{$\downarrow\!8.00$}}
& \maincell{57.02}{\textcolor{mainup}{$\uparrow\!1.03$}}
& \maincell{55.98}{\textcolor{mainup}{$\uparrow\!2.90$}}
& \maincell{46.67}{\textcolor{mainup}{$\uparrow\!9.17$}}
& \maincell{34.18}{\textcolor{maindown}{$\downarrow\!0.63$}}
& \maincell{37.78}{\textcolor{mainup}{$\uparrow\!0.56$}}
& \maincell{47.94}{\textcolor{mainup}{$\uparrow\!0.84$}} \\

Meta-Team
& \maincell{73.00}{\textcolor{mainup}{$\uparrow\!9.00$}}
& \maincell{64.82}{\textcolor{mainup}{$\uparrow\!8.83$}}
& \maincell{64.63}{\textcolor{mainup}{$\uparrow\!11.55$}}
& \maincell{\textbf{69.58}}{\textcolor{mainup}{$\uparrow\!32.08$}}
& \maincell{36.08}{\textcolor{mainup}{$\uparrow\!1.27$}}
& \maincell{\textbf{41.67}}{\textcolor{mainup}{$\uparrow\!4.45$}}
& \maincell{58.30}{\textcolor{mainup}{$\uparrow\!11.20$}} \\

\rowcolor{metateamblue}
\textbf{Ours}
& \maincell{\textbf{78.00}}{\textcolor{mainup}{$\uparrow\!14.00$}}
& \maincell{\textbf{67.32}}{\textcolor{mainup}{$\uparrow\!11.33$}}
& \maincell{\textbf{66.23}}{\textcolor{mainup}{$\uparrow\!13.15$}}
& \maincell{63.75}{\textcolor{mainup}{$\uparrow\!26.25$}}
& \maincell{\textbf{36.71}}{\textcolor{mainup}{$\uparrow\!1.90$}}
& \maincell{41.11}{\textcolor{mainup}{$\uparrow\!3.89$}}
& \maincell{\textbf{58.85}}{\textcolor{mainup}{$\uparrow\!11.75$}} \\

\midrule
\rowcolor{gray!12}
\multicolumn{8}{@{}l@{}}{\textbf{Mixed-Benchmark Evolution}} \\

Skill-MAS
& \maincell{36.00}{\textcolor{maindown}{$\downarrow\!28.00$}}
& \maincell{56.01}{\textcolor{mainup}{$\uparrow\!0.02$}}
& \maincell{57.83}{\textcolor{mainup}{$\uparrow\!4.75$}}
& \maincell{29.17}{\textcolor{maindown}{$\downarrow\!8.33$}}
& \maincell{35.44}{\textcolor{mainup}{$\uparrow\!0.63$}}
& \maincell{33.33}{\textcolor{maindown}{$\downarrow\!3.89$}}
& \maincell{41.30}{\textcolor{maindown}{$\downarrow\!5.80$}} \\

ReCreate
& \maincell{62.33}{\textcolor{maindown}{$\downarrow\!1.67$}}
& \maincell{59.76}{\textcolor{mainup}{$\uparrow\!3.77$}}
& \maincell{61.40}{\textcolor{mainup}{$\uparrow\!8.32$}}
& \maincell{40.83}{\textcolor{mainup}{$\uparrow\!3.33$}}
& \maincell{36.08}{\textcolor{mainup}{$\uparrow\!1.27$}}
& \maincell{36.11}{\textcolor{maindown}{$\downarrow\!1.11$}}
& \maincell{49.42}{\textcolor{mainup}{$\uparrow\!2.32$}} \\

Meta-Team
& \maincell{67.00}{\textcolor{mainup}{$\uparrow\!3.00$}}
& \maincell{64.90}{\textcolor{mainup}{$\uparrow\!8.91$}}
& \maincell{64.52}{\textcolor{mainup}{$\uparrow\!11.44$}}
& \maincell{58.33}{\textcolor{mainup}{$\uparrow\!20.83$}}
& \maincell{37.97}{\textcolor{mainup}{$\uparrow\!3.16$}}
& \maincell{40.56}{\textcolor{mainup}{$\uparrow\!3.34$}}
& \maincell{55.55}{\textcolor{mainup}{$\uparrow\!8.45$}} \\

\rowcolor{metateamblue}
\textbf{Ours}
& \maincell{\textbf{77.00}}{\textcolor{mainup}{$\uparrow\!13.00$}}
& \maincell{\textbf{65.55}}{\textcolor{mainup}{$\uparrow\!9.56$}}
& \maincell{\textbf{65.65}}{\textcolor{mainup}{$\uparrow\!12.57$}}
& \maincell{\textbf{67.08}}{\textcolor{mainup}{$\uparrow\!29.58$}}
& \maincell{\textbf{40.51}}{\textcolor{mainup}{$\uparrow\!5.70$}}
& \maincell{\textbf{41.67}}{\textcolor{mainup}{$\uparrow\!4.45$}}
& \maincell{\textbf{59.58}}{\textcolor{mainup}{$\uparrow\!12.48$}} \\

\bottomrule
\end{tabular}%
}
\label{tab:main_results}
\endgroup
\vspace{-6mm}
\end{table*}
\paragraph{Agent-level evolution.}
The Agent Reflector \(\Omega_A\) analyzes query \(q\) and the local trace \(\tau_{a_i}\) of each agent \(a_i\in E_q\) to update the agent's skill set \(S_{a_i}\):
\begin{equation}
    S_{a_i}' = \Omega_A(a_i,\tau_{a_i}, q,S_{a_i}),
\end{equation}
where \(S_{a_i}'\) is the resulting skill set.
The reflector examines the agent's reasoning steps, tool outputs, and intermediate artifacts for useful procedures or procedural errors.
It may add a useful procedure to the skill set or revise an existing skill to address an observed error.
It may retain the current set if the local trace offers no clear basis for either change.

After reflection, existing agent teams are updated in \(\mathcal{T}\), while new agent teams are added only after admission. 
Updated rule sets are retained for reuse within their agent teams, while updated skills are retained by the corresponding agents for reuse across agent teams. 
These updates guide team formation and collaborative execution on subsequent tasks with varying collaboration requirements.The algorithm is provided in the Appendix~\ref{appendix:algorithm}.


\section{Experiments}
\label{sec:experiments}
\vspace{-3mm}
In this section, we conduct experiments to answer four questions:
\textbf{Q1 (Overall Effectiveness):} How does \textsc{EvoTeam} perform against existing MAS methods across benchmarks under individual-benchmark and mixed-benchmark evolution?
\textbf{Q2 (Cross-Backbone Transfer):} Do evolved teams and interaction rules remain beneficial when evolution and evaluation LLMs differ?
\textbf{Q3 (Ablation Study):} How do team organization and interaction rules affect \textsc{EvoTeam}'s performance?
\textbf{Q4 (Cost Efficiency):} How does \textsc{EvoTeam} compare with existing MAS methods in performance and cost?

\begin{table*}[t]
\centering
\caption{\textbf{Same-backbone and cross-backbone evaluation.}
DeepSeek and GPT denote DeepSeek-V4.1-Flash and GPT-5.6-Luna,
respectively. Arrows indicate the evolution-to-evaluation direction.
Bold indicates the better score.}
\label{tab:backbone_transfer}
\vspace{-2mm}

\begingroup
\small
\setlength{\tabcolsep}{4pt}
\renewcommand{\arraystretch}{1.18}

\begin{tabular*}{\textwidth}{
    @{\extracolsep{\fill}}l*{9}{c}@{}
}
\toprule
\multirow{2}{*}{\textbf{Evolution $\rightarrow$ Evaluation}}
& \multicolumn{3}{c}{\textbf{GAIA}}
& \multicolumn{3}{c}{\textbf{LoCoBench Feat}}
& \multicolumn{3}{c}{\textbf{LoCoBench Refact}} \\
\cmidrule(lr){2-4}
\cmidrule(lr){5-7}
\cmidrule(lr){8-10}
& \textbf{MAS} & \textbf{Ours} & $\boldsymbol{\Delta}$
& \textbf{MAS} & \textbf{Ours} & $\boldsymbol{\Delta}$
& \textbf{MAS} & \textbf{Ours} & $\boldsymbol{\Delta}$ \\
\midrule

\rowcolor{gray!15}
\multicolumn{10}{@{}l@{}}{
    \textbf{Same-Backbone Evaluation}
} \\
DeepSeek $\rightarrow$ DeepSeek
& 58.00 & \textbf{80.00} & $+22.00$
& 66.32 & \textbf{68.93} & $+2.61$
& 65.15 & \textbf{68.16} & $+3.01$ \\
GPT $\rightarrow$ GPT
& 72.00 & \textbf{78.00} & $+6.00$
& 61.71 & \textbf{67.32} & $+5.61$
& 61.49 & \textbf{66.23} & $+4.74$ \\

\addlinespace[3pt]
\rowcolor{gray!15}
\multicolumn{10}{@{}l@{}}{
    \textbf{Cross-Backbone Evaluation}
} \\
DeepSeek $\rightarrow$ GPT
& 72.00 & \textbf{77.00} & $+5.00$
& 61.71 & \textbf{65.28} & $+3.57$
& 61.49 & \textbf{66.15} & $+4.66$ \\
GPT $\rightarrow$ DeepSeek
& 58.00 & \textbf{76.00} & $+18.00$
& 66.32 & \textbf{68.30} & $+1.98$
& 65.15 & \textbf{67.35} & $+2.20$ \\
\bottomrule
\end{tabular*}
\endgroup
\vspace{-6mm}
\end{table*}

\subsection{Experimental Setup}

\paragraph{Benchmarks.}
We use GAIA~\citep{mialon2024gaia} (multi-step web-based reasoning and retrieval), LoCoBench~\citep{qiu2025locobenchbenchmarklongcontextlarge} (long-context code implementation and cross-file refactoring), LOCA-Bench~\citep{zeng2026loca} (stateful application interaction via MCP), and BeyondSWE~\citep{chen2026beyondswecurrentcodeagent} (repository-level software maintenance).
Implementation details are in Appendix~\ref{appendix:dataset_details}.

\paragraph{Evaluation Metrics.}
We report GAIA answer accuracy, LOCA-Bench final-state success, BeyondSWE resolution rate (required tests pass without regressions), and LoCoBench Score (LCBS; functional correctness and software quality) on a common $[0,100]$ scale.
Detailed metric definitions are provided in Appendix~\ref{appendix:evaluation_metrics}.

\paragraph{Baselines.}
Baselines include hand-crafted SA (ReAct)~\citep{yao2023reactsynergizingreasoningacting}, non-evolutionary MAS, and OWL~\citep{hu2025owloptimizedworkforcelearning}; performance-driven search via AgentSquare~\citep{shang2025agentsquareautomaticllmagent}; and experience-driven ReCreate~\citep{hao2026recreatereasoningcreatingdomain}, Meta-Team~\citep{hao2026evolveteamcollaborativeselfevolution}, and Skill-MAS~\citep{lin2026skillmasevolvingmetaskillautomatic}.
Baseline details are provided in Appendix~\ref{appendix:baseline_details}.

\paragraph{Evolution and Evaluation Setup.}
Individual-benchmark evolution optimizes AgentSquare, the experience-driven baselines, and \textsc{EvoTeam} on each benchmark/subset's evolution split; SA, MAS, and OWL run without prior evolution.
Mixed-benchmark evolution uses a shuffled interleaved sequence combining all four evolution splits for the experience-driven baselines and \textsc{EvoTeam}, each using a consistent configuration.
Evaluation uses disjoint held-out splits.
By default, all experiments use GPT-5.6-Luna~\citep{openai2026gpt56}; avg@3 averages three independent rollout scores per test instance.
Detailed data partitions and evolution protocols are provided in Appendix~\ref{appendix:execution_evolution_procedure}.

\subsection{Overall Effectiveness (Q1)}
\label{sec:main-results}

\noindent\textbf{Obs. 1. \textsc{EvoTeam} achieves strong performance across benchmarks with different task requirements.}
Under individual-benchmark evolution, \textsc{EvoTeam} achieves the highest average score of 58.85 and leads on four of the six evaluation metrics.
Its advantage over Meta-Team, the strongest competing method by average score, spans GAIA, both LoCoBench tasks, and BeyondSWE DepMig.
The margin reaches five points (78.00 versus 73.00) on GAIA.
\textsc{EvoTeam} also outperforms the MAS baseline on every metric.
The breadth of these gains shows that the framework is effective across different collaboration requirements, highlighting the value of task-conditioned team and the interaction experience retained within them.

\begin{figure*}[t]
    \centering
    \begin{minipage}[t]{0.475\textwidth}
        \vspace{0pt}
        \centering
        \begin{minipage}[c][0.58\linewidth][c]{\linewidth}
            \centering
            \includegraphics[width=\linewidth]{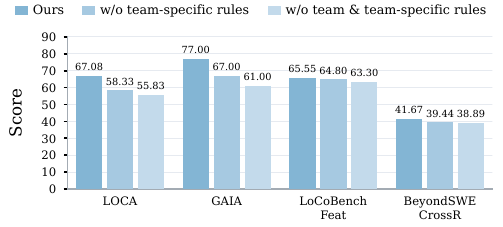}
        \end{minipage}\par
        \captionsetup{skip=4pt,justification=raggedright,singlelinecheck=false}
        \vspace{-2mm}
        \caption{Detailed component ablation results on four tasks under mixed-benchmark evolution.}
        \label{fig:ablation}
    \end{minipage}\hfill%
    \begin{minipage}[t]{0.475\textwidth}
        \vspace{0pt}
        \centering
        \begin{minipage}[c][0.58\linewidth][c]{\linewidth}
            \centering
            \includegraphics[width=\linewidth]{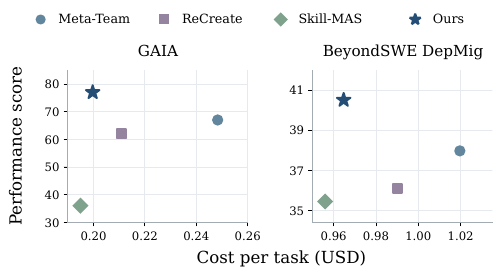}
        \end{minipage}\par
        \captionsetup{skip=4pt,justification=raggedright,singlelinecheck=false}
        \vspace{-2mm}
        \caption{Cost--performance on BeyondSWE DepMig and GAIA (mixed-benchmark evolution).}
        \label{fig:mixed_cost_performance}
    \end{minipage}
    \vspace{-6mm}
\end{figure*}

\noindent\textbf{Obs. 2. \textsc{EvoTeam} sustains its advantage when evolution combines heterogeneous benchmarks.}
Under the mixed-benchmark, \textsc{EvoTeam} ranks first on all six metrics and achieves an average score of 59.58, surpassing Meta-Team by 4.03 points.
Its lead is clear on GAIA and LOCA, while extending across the evaluation suite.
\textsc{EvoTeam}'s average score improves under the mixed-benchmark evolution relative to its individual-benchmark result.
Meta-Team's average score declines under the same comparison.
These results demonstrate \textsc{EvoTeam}'s advantage in reusing collaboration experience across the heterogeneous tasks with the different coordination requirements.

\subsection{Cross-Backbone Transfer (Q2)}
To address Q2, we examine whether \textsc{EvoTeam}'s evolved team and interaction rules can be reused across LLM backbones. We use DeepSeek-V4.1-Flash and GPT-5.6-Luna as the evolution and evaluation LLMs, and evaluate GAIA and the two LoCoBench tasks. For each configuration, we compare \textsc{EvoTeam} with the unevolved MAS baseline using the same evaluation LLM. Table~\ref{tab:backbone_transfer} reports the same- and cross-backbone evaluation results. We summarize two observations below.

\noindent\textbf{Obs. 3. Evolved collaboration transfers across backbones in both directions.}
When evolution and evaluation use different LLMs, \textsc{EvoTeam} consistently outperforms MAS in all six configurations.
On GAIA, both backbones benefit from artifacts evolved by the other model.
The gains reach 5.00 and 18.00 points for GPT-5.6-Luna and DeepSeek-V4.1-Flash respectively.
Both LoCoBench tasks also improve in each transfer direction, showing that the effect extends across task settings.
These bidirectional gains demonstrate that the collaboration experience accumulated through \textsc{EvoTeam}'s team-indexed evolution transfers effectively between the two evaluated LLM backbones.


\noindent\textbf{Obs. 4. Cross-backbone transfer retains performance close to the same-backbone setting.}
For each evaluation task, we hold the evaluation LLM fixed and vary only the LLM used to evolve the collaboration artifacts.
Across all six paired comparisons, cross-backbone performance remains within 4.00 points of the corresponding same-backbone performance.
On GAIA, cross-backbone transfer achieves 77.00 with GPT-5.6-Luna, compared with 78.00 under same-backbone evolution.
The two LoCoBench tasks also show similarly small performance gaps for both evaluation backbones.
This observed stability highlights \textsc{EvoTeam}'s robustness to changes in the evolution LLM and shows that its team-indexed collaboration experience remains effective across different backbone configurations.


\subsection{Ablation Study (Q3)}
\label{sec:ablation}

To address Q3, we compare the full \textsc{EvoTeam} with two variants: one removes learned interaction rules while retaining task-conditioned team, and other removes both components.
Figure~\ref{fig:ablation} reports their performance on four tasks.
We summarize two observations below. The complete ablation results are provided in the Appendix~\ref{appendix:component_ablation}.

\noindent\textbf{Obs. 5. Learned interaction rules strengthen collaboration within team.}
Across all four tasks, the full \textsc{EvoTeam} outperforms the variant that retains templates but removes interaction rules.
The largest gains appear on GAIA and LOCA, where the rules add 10.00 and 8.75 points.
The resulting performance gap supports the importance of retaining experience about how agents work together across successive task executions, beyond determining which agents participate.
The smaller gains on the coding tasks indicate that the contribution of this guidance varies across tasks.


\noindent\textbf{Obs. 6. Task-conditioned teams provide gains even without interaction rules.}
When interaction rules are absent from both variants, retaining templates improves performance on every task.
The largest gains appear on GAIA and LOCA, reaching 6.00 and 2.50 points respectively.
LoCoBench Feat and BeyondSWE also improve, extending this benefit to coding tasks.
Because neither variant uses learned interaction rules, this comparison highlights the value of task-conditioned team organization.
\textsc{EvoTeam}'s reusable team library supports selecting or creating a team structure suited to each specific task.
Together, the two ablations show the value of combining adaptable team structures with accumulated coordination experience.


\subsection{Cost Efficiency (Q4)}
\label{sec:cost_efficiency}

To address Q4, we compare \textsc{EvoTeam} with three MAS baselines in terms of task performance and cost per task under Mixed-Benchmark.
Figure~\ref{fig:mixed_cost_performance} presents the corresponding results on GAIA and BeyondSWE DepMig.
We summarize one main observation below.
The cost comparison for Individual-Benchmark is provided in the Appendix~\ref{sec:cost-analysis}.

\noindent\textbf{Obs. 7. \textsc{EvoTeam} achieves the strongest performance at competitive cost.}
On both evaluated tasks, \textsc{EvoTeam} achieves the highest score while costing less per task than Meta-Team and ReCreate.
Skill-MAS has lower cost but also lower overall performance on both tasks.
These direct comparisons place \textsc{EvoTeam} in a favorable position on the performance--cost frontier.
The result further highlights the practical value of its team-scoped collaboration: evolved team structures and learned interaction rules improve task outcomes while keeping average per-task cost competitive.
\section{Related Work}
\label{sec:related_work}

\paragraph{Individual Agent Evolution.}
Experience-driven approaches improve agent capabilities
through scaffold revision, skill acquisition, and
cross-task experience retrieval~\citep{hao2026recreatereasoningcreatingdomain,zheng2025skillweaverwebagentsselfimprove,li2025crosstaskexperientiallearningllmbased}.
By retaining reusable knowledge from previous task executions, agents
progressively improve their reasoning and tool use.
However, individual improvement does not necessarily
resolve problems in responsibility assignment and
information exchange during collaboration.

\paragraph{Collaborative Multi-Agent Evolution.}
Meta-Team uses collective execution experience to evolve
agent behaviors, inter-agent coordination, and team
organization~\citep{hao2026evolveteamcollaborativeselfevolution}.
Skill-MAS refines orchestration knowledge through rollout
and reflection, while EMAS validates prompt and topology
revisions using execution evidence~\citep{lin2026skillmasevolvingmetaskillautomatic,fei2026emasstabilizingmultiagentevolution}.
Our work focuses on sustaining collaboration evolution
across heterogeneous tasks.
\textsc{EvoTeam} maintains and evolves reusable teams
with separate interaction rules, allowing coordination
experience to accumulate within compatible team contexts
while individual skills remain reusable across teams.
Further details are provided in Appendix~\ref{appendix:related_work}.
\section{Conclusion}
\label{sec:conclusion}
In this work, we presented \textsc{EvoTeam}, a framework for continual collaboration evolution across heterogeneous tasks. It maintains a library of reusable teams and jointly evolves their organization, team-specific interaction rules, and individual agent skills, accumulating coordination experience within compatible team contexts while preserving skill reuse across teams. Across four benchmarks and six task categories, \textsc{EvoTeam} leads all six metrics under mixed-benchmark evolution, averaging 59.58 versus 55.55 for the strongest baseline. Cross-backbone evaluations and ablations further support the transferability of evolved collaboration experience and the value of jointly evolving team organization and coordination practices.
\clearpage


\section*{Limitations}
Our evaluation does not systematically examine collaborative
evolution in multimodal interactive environments.
In such settings, agents need to interpret visual observations,
communicate relevant evidence, and coordinate actions as
the environment changes.
Further investigation is needed into how multimodal execution
traces can guide the refinement of team organization,
interaction rules, and agent skills.
Future work could extend \textsc{EvoTeam} to these settings
and evaluate how accumulated collaboration experience
supports tasks involving visual perception and interaction.
Additionally, \textsc{EvoTeam} accumulates experience
through explicit updates while keeping the
LLM parameters fixed.
The effective use of this experience depends
on the base model's ability to correctly interpret and
apply the stored guidance.
Combining experience reuse with model adaptation could
help internalize frequently useful coordination practices.
\section*{Ethical Considerations}

\textsc{EvoTeam} aims to improve multi-agent collaboration through
reusable teams and accumulated execution experience.
Like other LLM-based agent systems, it may inherit biases from
underlying models and produce inaccurate outputs or unintended
tool actions.
Persistent experience accumulation also requires attention to
sensitive information retained in execution traces and reusable
knowledge.
Applications beyond the evaluated benchmarks should therefore
incorporate appropriate access controls, privacy protection,
and human oversight for consequential actions.
Accumulated interaction rules and agent skills should be
reviewed for their suitability before reuse in new contexts.
The use of underlying models, datasets, and tools should
respect their respective licenses and usage policies.

\bibliography{main}

\clearpage
\appendix

\definecolor{ticerPromptBlue}{HTML}{426B95}
\definecolor{ticerPromptGray}{HTML}{F3F4F6}
\tcbset{ticerAppendixBox/.style={
    enhanced,
    width=\linewidth,
    colback=ticerPromptGray,
    colframe=ticerPromptBlue,
    colbacktitle=ticerPromptBlue,
    coltitle=white,
    fonttitle=\fontsize{10}{11.5}\selectfont\bfseries\sffamily,
    fontupper=\fontsize{10}{11.5}\selectfont,
    boxrule=0.5pt,
    arc=1mm,
    left=6pt, right=6pt,
    top=5pt, bottom=5pt,
    toptitle=3pt, bottomtitle=3pt,
    before skip=7pt plus 1pt minus 1pt,
    after skip=7pt plus 1pt minus 1pt,
    before upper={\setlength{\parindent}{0pt}%
        \setlength{\parskip}{3pt plus 0.5pt minus 0.5pt}%
        \setlength{\emergencystretch}{1em}%
        \clubpenalty=10000 \widowpenalty=10000 \brokenpenalty=10000}
}}
\newenvironment{ticerItemize}{%
    \begin{list}{\textbullet}{%
        \setlength{\leftmargin}{1.2em}%
        \setlength{\labelwidth}{0.7em}%
        \setlength{\labelsep}{0.5em}%
        \setlength{\topsep}{3pt}%
        \setlength{\partopsep}{0pt}%
        \setlength{\itemsep}{3pt}%
        \setlength{\parsep}{0pt}}%
}{\end{list}}
\newcounter{ticerPromptStep}
\newenvironment{ticerEnumerate}{%
    \interlinepenalty=10000 
    \begin{list}{\arabic{ticerPromptStep}.}{%
        \usecounter{ticerPromptStep}%
        \setlength{\leftmargin}{1.5em}%
        \setlength{\labelwidth}{1em}%
        \setlength{\labelsep}{0.5em}%
        \setlength{\topsep}{3pt}%
        \setlength{\partopsep}{0pt}%
        \setlength{\itemsep}{2pt}%
        \setlength{\parsep}{0pt}}%
}{\end{list}}


\makeatletter
\if@twocolumn\else\twocolumn\fi
\@ifpackageloaded{lineno}{\switchlinenumbers\setlength{\linenumbersep}{1.6cm}}{}
\makeatother
\setlength{\columnsep}{0.6cm} 
\flushbottom

\setcounter{dbltopnumber}{1}
\renewcommand{\dbltopfraction}{0.85}
\renewcommand{\dblfloatpagefraction}{0.95}
\raggedbottom
\section{Algorithms}
\label{appendix:algorithm}

Algorithm~\ref{alg:team_evolution} summarizes the sequential
evolution procedure of \textsc{EvoTeam}, following the notation
in Section~\ref{sec:method}.
The framework maintains a global agent pool $\mathcal{A}$,
a library of reusable agent teams $\mathcal{T}$, and
team-specific interaction-rule sets
$\{\mathcal{R}_T\}_{T\in\mathcal{T}}$.
For each evolution query, it selects a compatible agent team
or creates a provisional one with an empty rule set.
The Chairman recruits an execution team from its candidate
roster and assigns subtasks.
The selected team's rules are injected into the recruited
agents' contexts as soft coordination guidance.

\textsc{Execute} includes subtask execution using accumulated
agent skills, peer coordination, native task evaluation,
and trace collection.
Agents select applicable rules according to their descriptions
and current coordination requirements.
The Chairman monitors progress and adjusts $\pi_q$ by
reassigning subtasks within $E_q$ when needed.
The resulting composite trace contains individual agent
traces and the corresponding inter-agent handoff trace.

After execution, the three reflectors update the agent team,
its interaction rules, and individual agent skills.
Team reflection revises the candidate roster when needed
and updates the team description accordingly.
For a provisional team, the same reflection pass determines
admission based on whether its roster supports the required
division of work and its organization is reusable.
Existing teams retain their identities when revised.
Rules are retained with their associated teams, while
updated skills remain with the corresponding agents
regardless of provisional-team admission.
After the final query, all persistent updates are frozen
for evaluation on the disjoint holdout split.

\begin{algorithm}[t]
\small
\caption{Sequential Collaborative Team Evolution}
\label{alg:team_evolution}
\begin{algorithmic}[1]
\REQUIRE Evolution queries $(q_1,\ldots,q_K)$;
    agent pool $\mathcal{A}$;
    team library $\mathcal{T}$;
    rule sets $\{\mathcal{R}_T\}_{T\in\mathcal{T}}$
\ENSURE Frozen agent pool, team library, and rule sets

\FOR{each $q$ in $(q_1,\ldots,q_K)$}
    \STATE $T_q \gets \Phi_S(q,\mathcal{T})$
    \IF{$T_q=\emptyset$}
        \STATE $\tilde{T}_q \gets \Phi_C(q,\mathcal{A})$
        \STATE $T_q \gets \tilde{T}_q$,
            \quad $\mathcal{R}_{T_q}\gets\emptyset$
    \ENDIF

    \STATE $E_q \gets \Phi_{\text{Chair}}(q,T_q)$
    \STATE $(U_q,\pi_q) \gets
        \Phi_{\text{Chair}}(q,T_q,E_q)$
    \STATE Inject $\mathcal{R}_{T_q}$ into agents' contexts
        as soft guidance
    \STATE $\tau_c \gets
        \textsc{Execute}(q,T_q,E_q,U_q,\pi_q)$
    \STATE $\langle\{\tau_{a_i}\}_{a_i\in E_q},\tau_t\rangle
        \gets \tau_c$

    \STATE $T_q' \gets \Omega_T(T_q,\tau_c,q)$
    \STATE $\mathcal{R}_{T_q}' \gets
        \Omega_R(\mathcal{R}_{T_q},\tau_t,q)$
    \FOR{each $a_i\in E_q$}
        \STATE $S_{a_i}' \gets
            \Omega_A(a_i,\tau_{a_i},q,S_{a_i})$
        \STATE $S_{a_i} \gets S_{a_i}'$
            \COMMENT{persist in $\mathcal{A}$}
    \ENDFOR

    \IF{$T_q\in\mathcal{T}$}
        \STATE Revise $T_q$ to $T_q'$ in $\mathcal{T}$,
            preserving its identity
        \STATE Save $\mathcal{R}_{T_q}'$ with the revised team
    \ELSIF{$T_q'$ is admitted by $\Omega_T$}
        \STATE Add $T_q'$ to $\mathcal{T}$
        \STATE Save $\mathcal{R}_{T_q}'$ with the admitted team
    \ENDIF
\ENDFOR

\STATE Freeze $\mathcal{A}$, $\mathcal{T}$, and
    $\{\mathcal{R}_T\}_{T\in\mathcal{T}}$
\RETURN $\mathcal{A}$, $\mathcal{T}$, and
    $\{\mathcal{R}_T\}_{T\in\mathcal{T}}$
\end{algorithmic}
\end{algorithm}

\section{Experimental Details}
\label{app:experimental_details}

We describe the benchmarks, data splits, native evaluation metrics, and
execution settings used in our experiments. Each setting uses a designated
evolution split and a disjoint holdout split. Table~\ref{tab:appendix_experimental_settings}
summarizes splits and native evaluation protocols.

\subsection{Benchmark Details}
\label{appendix:dataset_details}

Our evaluation covers four task settings: multi-step information seeking
(GAIA), long-context software development (LoCoBench), tool-based productivity
workflows (LOCA-Bench), and repository-level software engineering (BeyondSWE).
We use the same subsets and evolution/holdout splits as
Meta-Team~\citep{hao2026evolveteamcollaborativeselfevolution}, and additionally combine the four
evolution splits together for mixed-benchmark evolution to effectively evaluate a single shared evolving system on the heterogeneous tasks.
\par\smallskip
\noindent\textbf{(i) GAIA.}~\citep{mialon2024gaia} evaluates general-purpose assistants on real-world
questions requiring reasoning, web browsing, tool use, and, for some questions,
multimodal understanding. Questions are organized into three difficulty levels
and have unambiguous reference answers; some include supplementary files.
Following Meta-Team, we use the 120-instance validation subset, with 20 instances for evolution and 100 disjoint instances for holdout
evaluation. The evolution split contains 6, 9, and 5 instances from Levels 1,
2, and 3, respectively; the holdout split contains 38, 49, and 13 instances.
Agents can read and write files, run shell commands, search online, and
fetch webpages.
\par\smallskip
\noindent\textbf{(ii) LoCoBench.}~\citep{qiu2025locobenchbenchmarklongcontextlarge} evaluates long-context software
development through 8,000 scenarios across 10 programming languages and
eight task categories, with contexts of 10K to 1M tokens that require
reasoning across multiple files. We use the Python Feature Implementation (FI)
and Cross-File Refactoring (CR) categories. FI requires adding functionality to
an existing codebase, while CR requires restructuring code across files while
preserving its behavior. Each category contains 100 instances in our
experiments: 20 for evolution and 80 for disjoint holdout evaluation, with each instance executed in its own isolated workspace.
\par\smallskip
\noindent\textbf{(iii) LOCA-Bench.}~\citep{zeng2026loca} evaluates language agents under controllable
context growth by scaling environment states while preserving task semantics.
The full benchmark contains 15 task types, seven environment-description
lengths (8K, 16K, 32K, 64K, 96K, 128K, and 256K), and five
environment seeds. Environment-description length measures
the tokenized information available through the task environment's tool
outputs; it is distinct from the model's context-window limit.
We select eight task types spanning e-commerce, academic, and education
workflows, summarized in Table~\ref{tab:loca_task_types}.
Evolution uses these eight tasks at 96K with seeds 101 and 102, giving
$8\times2=16$ cases. Holdout evaluation uses seeds
$\{42,123,456,789,2024\}$ at 8K, 16K, 32K, 64K, 128K, and 256K,
giving $8\times5\times6=240$ cases. The evolution and holdout configurations
are disjoint in both environment seeds and description lengths, testing
cross-seed and cross-scale generalization for the selected task types.
Each case runs with isolated local mock MCP services and is scored by the corresponding
native environment-state evaluator.

\begin{table*}[t]
\centering
\caption{The eight LOCA-Bench task types used in our experiments, following
Meta-Team~\citep{hao2026evolveteamcollaborativeselfevolution}. Each task type has a total of 30 distinct holdout cases
(five seeds across six environment-description lengths).}
\label{tab:loca_task_types}
\small
\setlength{\tabcolsep}{5pt}
\renewcommand{\arraystretch}{1.12}
\begin{tabularx}{\textwidth}{@{}
    >{\raggedright\arraybackslash}p{0.30\textwidth}
    >{\centering\arraybackslash}p{0.09\textwidth}
    >{\raggedright\arraybackslash}p{0.23\textwidth}
    >{\raggedright\arraybackslash}X@{}}
\toprule
\textbf{Task} & \textbf{Holdout} & \textbf{Domain} & \textbf{Key MCP services} \\
\midrule
WoocommerceNewWelcome & 30 & E-commerce CRM & woocommerce, google\_cloud, email \\
WoocommerceStockAlert & 30 & E-commerce purchasing & woocommerce, google\_sheet, email \\
FilterLowSellingProducts & 30 & E-commerce clearance & woocommerce, email \\
ApplyPhDEmail & 30 & Academic & email, pdf\_tools \\
SetConfCrDdl & 30 & Academic & calendar, email \\
CourseAssistant & 30 & Education & excel, email \\
CanvasArrangeExam & 30 & Education & canvas, email, excel \\
CanvasListTest & 30 & Education & canvas \\
\midrule
\textbf{Total} & \textbf{240} & & \\
\bottomrule
\end{tabularx}
\end{table*}
\par\smallskip
\noindent\textbf{(iv) BeyondSWE.}~\citep{chen2026beyondswecurrentcodeagent} evaluates software engineering beyond
single-repository bug fixing. Its 500 instances are drawn from 246 real-world
GitHub repositories and cover cross-repository issue resolution, domain-specific
fixes, dependency migration, and repository generation from specifications.
We use the CrossRepo and DepMigrate tracks. CrossRepo requires reasoning with
information from related external repositories, while DepMigrate requires
repository-wide changes to accommodate breaking dependency updates.
Following Meta-Team, we randomly select 20 instances per track for evolution
and hold out the rest: CrossRepo has 200 instances (20 evolution, 180 holdout)
and DepMigrate has 178 (20 evolution, 158 holdout). Each track is evolved
separately and evaluated on its own holdout split.
Each instance runs in an isolated repository environment, and submitted changes
are evaluated with native Fail-to-Pass (F2P) and Pass-to-Pass (P2P) tests.
\par\smallskip
\noindent\textbf{(v) Mixed-Benchmark.} combines the four source benchmarks into a single mixed-task setting.
Its evolution split contains 116 instances:
20 GAIA, 20 LoCoBench FI, 20 LoCoBench CR, 16 LOCA-Bench, 20 BeyondSWE
CrossRepo, and 20 BeyondSWE DepMigrate. These instances are materialized once
in a fixed global-shuffle order using seed 20260828. 
A shared system evolves sequentially over the mixed task stream and is frozen
before evaluating the holdout groups. Each task retains its source benchmark's
tools, environment, output contract, and evaluator. Results are reported using the respective source-native evaluation metrics for each of the corresponding holdout groups.

\begin{table*}[t]

\centering
\caption{Data splits and evaluation protocols. Counts denote
distinct instances or environment configurations.}
\label{tab:appendix_experimental_settings}
\small
\setlength{\tabcolsep}{4pt}
\renewcommand{\arraystretch}{1.12}
\begin{tabularx}{\textwidth}{
    >{\raggedright\arraybackslash}p{0.24\textwidth}
    >{\centering\arraybackslash}p{0.10\textwidth}
    >{\centering\arraybackslash}p{0.10\textwidth}
    >{\raggedright\arraybackslash}X
    >{\centering\arraybackslash}p{0.18\textwidth}}
\toprule
\textbf{Benchmark / track} & \textbf{Evolution} & \textbf{Holdout}
& \textbf{Native environment} & \textbf{Metric} \\
\midrule
GAIA & 20 & 100 & Attachments and web tools & Pass rate \\
LoCoBench FI & 20 & 80 & Isolated code workspace & LCBS (0--100) \\
LoCoBench CR & 20 & 80 & Isolated code workspace & LCBS (0--100) \\
LOCA-Bench & 16 & 240 & Local mock MCP services & Pass rate \\
BeyondSWE CrossRepo & 20 & 180 & Isolated repository & Pass rate \\
BeyondSWE DepMigrate & 20 & 158 & Isolated repository & Pass rate \\
mixed-benchmark & 116 & 838 & Source-specific environments & Source-native metrics \\
\bottomrule
\end{tabularx}
\end{table*}

\subsection{Evaluation Metrics}
\label{appendix:evaluation_metrics}

We follow each benchmark's native evaluation protocol.
All aggregate scores are reported on a $[0,100]$ scale,
with higher values indicating better overall performance across different benchmarks.

\noindent\textbf{(i) GAIA.}
We report answer accuracy using the official
evaluator~\citep{mialon2024gaia}.
A prediction is correct only if the complete final answer
matches the reference under the evaluator's normalization rules.
These rules handle differences in capitalization, whitespace,
punctuation, and numeric formatting as appropriate to the
answer type.
For list-valued answers, the number of elements and all
corresponding elements must match.
Both semantically similar but unmatched answers and partially correct answers are scored as incorrect.

\noindent\textbf{(ii) LOCA-Bench.}
We report task success rate under the official execution-based
evaluation protocol~\citep{zeng2026loca}.
Task-specific evaluation scripts check whether the final
environment state satisfies the required outcomes, such as
correctly updated records or generated artifacts.
Each run receives a binary score: one if the task passes
the evaluator and zero otherwise.
Success therefore depends on the resulting environment state,
rather than the agent's mere textual claim of completion.

\noindent\textbf{(iii) BeyondSWE.}
For CrossRepo and DepMigrate, we report the percentage of
resolved instances~\citep{chen2026beyondswecurrentcodeagent}.
An instance is resolved only when the submitted changes pass
all required Fail-to-Pass (F2P) tests while preserving all
Pass-to-Pass (P2P) tests.
F2P tests verify that the target issue has been addressed,
whereas P2P tests check that previously working functionality
remains intact.
Passing only a subset of the specified required tests cannot be counted as a resolved instance.

\noindent\textbf{(iv) LoCoBench.}
For Feature Implementation and Cross-File Refactoring,
we specifically report the LoCoBench Score
(LCBS)~\citep{qiu2025locobenchbenchmarklongcontextlarge},
which combines four normalized dimensions.

Software Engineering Excellence (SE)
assesses architectural consistency, dependency handling,
cross-file reasoning, robustness, and solution completeness.
Functional Correctness (FC)
assesses compilation, unit and integration test performance,
and incremental development capability.
Code Quality (CQ)
assesses security, maintainability, and coding-style adherence.
Long-Context Utilization (LCU)
assesses task-relevant information use in long contexts and effective cross-session retention.

Each dimension averages its constituent normalized metrics and lies in $[0,1]$.
For each instance, the score on our reporting scale is given by

\begin{equation}
\begin{aligned}
\mathrm{LCBS}_{100}
=&100\bigl(0.4\,SE+0.3\,FC\\
         &+0.2\,CQ+0.1\,LCU ).
\end{aligned}
\end{equation}
This is equivalent to multiplying the native $[0,5]$ LCBS
by 20. We average per-instance scores within each evaluated
subset. Unlike the binary metrics above, LCBS assigns
continuous scores that reflect performance across the
four dimensions.

\subsection{Baseline Details}
\label{appendix:baseline_details}
We compare our method with the following seven baselines.
All these methods use GPT-5.6-Luna for task execution and, where applicable,
for reflection and optimization. They share the same benchmark-specific
tool access, resource budgets, and evolution/holdout splits as \textsc{EvoTeam}.

\noindent\textbf{(i) SA} uses a single agent with a ReAct-style
workflow~\citep{yao2023reactsynergizingreasoningacting}, alternating between reasoning and tool use to solve
each given task, without involving any form of experience-driven evolution.

\noindent\textbf{(ii) MAS} uses a multi-agent team with the same collaboration
structure as ours. The team configuration remains unchanged across
tasks, with no experience-driven evolution. MAS is evaluated
directly and independently on the holdout split.

\noindent\textbf{(iii) OWL}
~\citep{hu2025owloptimizedworkforcelearning} organizes task execution through a
Planner--Coordinator--Worker hierarchy.
We use the official implementation with its original structure under the
common backbone, tool-access, and budget settings. OWL does not perform
experience-driven evolution and is evaluated directly on the corresponding holdout split.

\noindent\textbf{(iv) AgentSquare}~\citep{shang2025agentsquareautomaticllmagent} searches a modular agent
design space comprising planning, reasoning, tool use, and memory.
We use its official implementation and released search configuration,
with the candidate scaffold pool adopted from Meta-Team.
Search is conducted on the designated evolution split, and the selected
configuration is kept fixed for the subsequent holdout evaluation.

\noindent\textbf{(v) Skill-MAS}
~\citep{lin2026skillmasevolvingmetaskillautomatic} evolves a meta-skill that guides
task decomposition, agent construction, and workflow orchestration.
It combines multiple trajectory rollouts with selective reflection
to distill reusable orchestration principles from execution experience.
We use the official implementation with $k=2$, sampling two
separate trajectories per task in each evolution round.
Evolution uses only the designated evolution split, and the evolved
meta-skill is frozen before holdout evaluation.

\noindent\textbf{(vi) ReCreate}~\citep{hao2026recreatereasoningcreatingdomain} is an experience-driven framework
for optimizing agent scaffolds.
Its agent-as-optimizer mechanism inspects execution trajectories,
submitted artifacts, and evaluation feedback to propose changes
to prompts, tools, and memory.
Instance-level proposals are synthesized into reusable
updates to the shared scaffold.
We use the official implementation under the common model,
tool-access, and budget settings.
Evolution uses only the designated evolution split, and the final
scaffold is frozen for holdout evaluation.
 
\noindent\textbf{(vii) Meta-Team}~\citep{hao2026evolveteamcollaborativeselfevolution} improves multi-agent systems
through collaborative analysis of execution experience.
Agents retain their local execution contexts and exchange relevant
findings to support updates to individual agent behavior,
inter-agent collaboration, and team organization.
We use its official implementation and apply its experience-driven
evolution procedure specifically on the predefined designated evolution split.
\subsection{Evolution and Evaluation}
\label{appendix:execution_evolution_procedure}

\paragraph{Individual-Benchmark Setting.}
Each benchmark or track is treated as an independent
evolution and evaluation setting.
SA, MAS, and OWL are evaluated directly on the holdout
split without prior evolution.
AgentSquare, ReCreate, Meta-Team, Skill-MAS, and
\textsc{EvoTeam} are optimized on the corresponding
evolution split using their respective optimization procedures and initial configurations.
No accumulated experience is shared across benchmarks
or tracks.
The configurations are frozen before evaluation on the disjoint holdout split.

\paragraph{Mixed-Benchmark Setting.}
ReCreate, Meta-Team, Skill-MAS, and \textsc{EvoTeam}
participate in mixed-benchmark evolution.
The evolution stream contains 116 instances:
20 GAIA, 20 LoCoBench FI, 20 LoCoBench CR,
16 LOCA-Bench, 20 BeyondSWE CrossRepo, and
20 BeyondSWE DepMigrate.
These instances are shuffled once using seed 20260828,
and the order is fixed across methods.
All four methods start from the same shared initial
agent pool.
Each method evolves a single system over the mixed task stream using its own update procedure, carrying experience across benchmarks.
Each task retains its source-specific tools, environment, output contract, and evaluator. 
The final system is frozen and evaluated separately on all source holdout groups using their native metrics.
\paragraph{Resource budgets.}
Table~\ref{tab:appendix_runtime_budgets} lists the per-case time, message, and
cost limits. For LOCA-Bench, these limits remain consistently fixed across all considered different environment-description lengths.
\paragraph{Implementation Details.}
Unless otherwise specified, all methods use
GPT-5.6-Luna~\citep{openai2026gpt56} for task execution and, where
applicable, reflection and optimization, with a
temperature of 0.2 and a maximum output length of
32,768 tokens.
Within each evaluation setting, methods share the same
benchmark-specific tool access, per-task resource limits,
and corresponding data partitions.

\begin{table}[t]
\centering
\caption{Specified per-case resource budgets for both evolution and holdout evaluation phases.}
\label{tab:appendix_runtime_budgets}

\small
\setlength{\tabcolsep}{3pt}
\renewcommand{\arraystretch}{1.12}

\begin{tabularx}{\columnwidth}{
    @{}l
    *{3}{>{\centering\arraybackslash}X}@{}
}
\toprule
\textbf{Benchmark}
& \shortstack{\textbf{Max. time}\\\textbf{(s)}}
& \shortstack{\textbf{Max.}\\\textbf{messages}}
& \shortstack{\textbf{Max. cost}\\\textbf{(\$)}} \\
\midrule
GAIA       & 900  & 300 & 2.40  \\
LoCoBench  & 1800 & 400 & 2.40  \\
BeyondSWE  & 1800 & 600 & 2.40  \\
LOCA-Bench & 3200 & 600 & 12.00 \\
\bottomrule
\end{tabularx}
\end{table}



\section{Additional Experiments}
\label{appendix:additional_experiments}

\begin{figure*}[t]
    \centering
    \includegraphics[width=\textwidth]
    {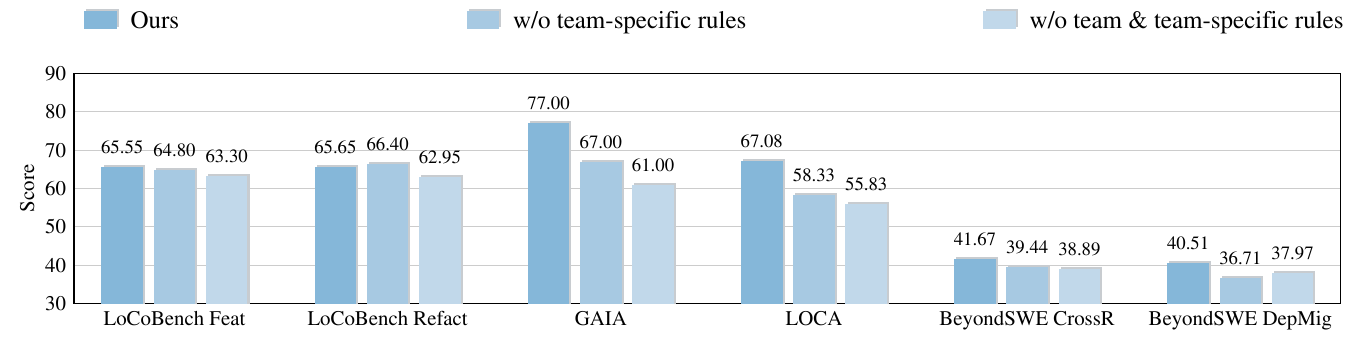}
    \caption{Ablation results across six different evaluation metrics.
    We compare the proposed full model with variants that remove team-specific handoff rules, or both reusable agent teams and handoff rules.}
    \label{fig:appendix_ablation}
\par\vspace{\floatsep}
\centering
    \includegraphics[width=\textwidth]{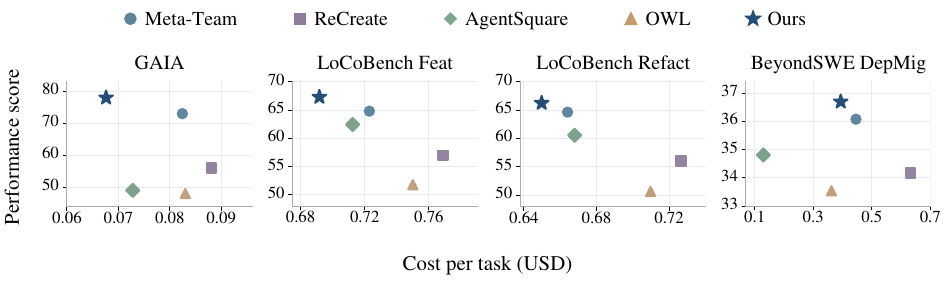}
    \caption{Performance versus estimated cost per task
    under Individual-Benchmark evaluation.
    Higher scores and lower costs are both preferred in this comparison.}
    \label{fig:cost}
\end{figure*}

We complement the main results with analyses across
GAIA difficulty levels, LoCoBench evaluation dimensions,
and component ablations across all the evaluated benchmarks
and task subsets.

\subsection{Component Ablation}
\label{appendix:component_ablation}

Figure~\ref{fig:appendix_ablation} reports the
ablation across six benchmark settings.
We compare the full framework with two variants:
removing team-specific rules, and removing both teams
and team-specific rules.




\paragraph{Obs. 8: The complete framework performs best in five of six settings.}
The full framework also outperforms the variant without
both reusable agent teams and interaction rules in all six settings.
On GAIA, LOCA, LoCoBench Feat, and BeyondSWE CrossR,
removing the two components lowers performance,
supporting the value of combining team organization with
learned coordination guidance.
The remaining settings show two exceptions:
removing rules slightly improves LoCoBench Refact
(66.40 vs.\ 65.65), while removing both components
outperforms removing rules alone on BeyondSWE DepMig
(37.97 vs.\ 36.71).
Despite these exceptions, the complete design provides
the strongest overall results across the evaluated tasks.

\subsection{Cost Analysis}
\label{sec:cost-analysis}

Figure~\ref{fig:cost} compares both
performance and cost per task under individual-benchmark
evolution.
\paragraph{Obs. 9: Collaboration reuse offers a favorable performance--cost trade-off.}
Figure~\ref{fig:cost} shows that \textsc{EvoTeam} achieves the
highest score at the lowest estimated cost on GAIA and both
LoCoBench tasks.
On BeyondSWE DepMig, it outperforms Meta-Team and ReCreate
at a lower cost; AgentSquare and OWL are cheaper but score lower.
This trade-off is consistent with the reuse mechanisms:
reusable agent teams provide established divisions of work, while
interaction rules supply relevant coordination guidance.
Reusing both reduces the need to establish same collaboration
procedures anew for tasks with compatible requirements.

\setcounter{topnumber}{1}
\begin{table*}[t]
\centering
\caption{LoCoBench dimension scores after mixed-benchmark evolution.
The highest score for each dimension within each subset is in \textbf{bold}.}
\label{tab:locobench_dimension_breakdown}
\small
\setlength{\tabcolsep}{6pt}
\renewcommand{\arraystretch}{1.15}

\begin{tabular*}{\textwidth}{@{\extracolsep{\fill}}llrrrr@{}}
\toprule
\textbf{Subset} & \textbf{Method}
& \textbf{SE} & \textbf{FC} & \textbf{CQ} & \textbf{LCU} \\
\midrule

Feat. & Skill-MAS
& 0.4916 & 0.5744 & 0.5878 & 0.7360 \\
& ReCreate
& 0.3766 & 0.6832 & \textbf{0.8407} & 0.7384 \\
& Meta-Team
& 0.4923 & 0.7231 & 0.7802 & \textbf{0.7915} \\
& Ours
& \textbf{0.4979} & \textbf{0.7239} & 0.8064 & 0.7789 \\

\midrule

Refact. & Skill-MAS
& 0.5109 & 0.6264 & 0.5960 & 0.6686 \\
& ReCreate
& 0.4375 & 0.7181 & \textbf{0.7676} & 0.7001 \\
& Meta-Team
& 0.5204 & 0.7148 & 0.7594 & 0.7068 \\
& Ours
& \textbf{0.5273} & \textbf{0.7494} & 0.7428 & \textbf{0.7219} \\

\bottomrule
\end{tabular*}

\raggedright
SE: software engineering excellence; FC: functional correctness;
CQ: code quality; LCU: long-context utilization.
\end{table*}

\begin{table}[t]
\centering
\caption{GAIA pass rates (\%) by difficulty level. Scores use all instances
in each level. The highest score in each row is in
\textbf{bold}.}
\label{tab:gaia_level_breakdown}
\small
\setlength{\tabcolsep}{3pt}
\renewcommand{\arraystretch}{1.15}
\begin{tabular*}{\linewidth}{@{\extracolsep{\fill}}lcccc@{}}
\toprule
\textbf{Level} & \textbf{Skill-MAS} & \textbf{Meta-Team}
& \textbf{ReCreate} & \textbf{Ours} \\
\midrule
Level 1 & 42.11 & 78.95 & 69.30 & \textbf{81.58} \\
Level 2 & 32.65 & 67.35 & 65.31 & \textbf{83.67} \\
Level 3 & 30.77 & 30.77 & 30.77 & \textbf{38.46} \\
\bottomrule
\end{tabular*}
\end{table}

\subsection{GAIA Difficulty Levels}
\label{appendix:gaia_level_results}
\label{appendix:benchmark_breakdown}

Table~\ref{tab:gaia_level_breakdown} reports GAIA pass rates
by difficulty level under the Mixed-Benchmark setting.
Evolved methods are evaluated on the GAIA holdout split.

\paragraph{Obs. 10: Gains extend across GAIA difficulty levels.}
\textsc{EvoTeam} achieves the highest pass rate at all three levels.
Its gains over Meta-Team are 2.63, 16.32, and 7.69 percentage
points on Levels~1, 2, and~3, respectively.
The largest gain occurs on Level~2, showing that the benefits
of mixed-benchmark evolution extend beyond the easiest tasks.
However, Level~3 scores remain much lower for all methods,
indicating that the hardest tasks continue to pose substantial challenges.

\subsection{LoCoBench Dimensions}
\label{appendix:locobench_dimension_results}

\begin{figure}[t]
\centering
    \includegraphics[width=\linewidth]{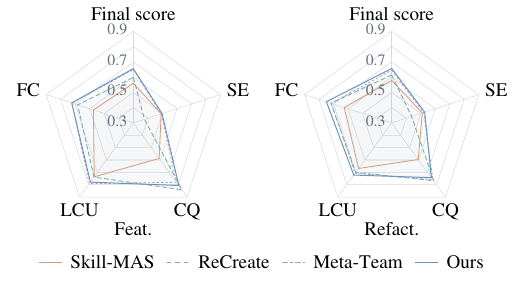}
    \caption{Fine-grained performance on the LoCoBench
    Feature Implementation (Feat.) and Cross-File Refactoring
    (Refact.) subsets.
     Higher is better.}
    \label{fig:appendix_locobench_radar}
\end{figure}

Table~\ref{tab:locobench_dimension_breakdown} and Figure~\ref{fig:appendix_locobench_radar} reports
dimension-level results under the mixed-benchmark setting,
evaluated on the LoCoBench Feature Implementation
(Feat.) and Cross-File Refactoring (Refact.) holdout
splits after mixed-benchmark evolution.
We report software engineering excellence (SE),
functional correctness (FC), code quality (CQ), and
long-context utilization (LCU).
This breakdown shows how each dimension contributes to overall performance.
\suppressfloats[t]

\paragraph{Obs. 11: LoCoBench gains are strongest in SE and FC.}
\textsc{EvoTeam} leads in SE and FC on both subsets, although
its margins over Meta-Team on Feat.\ are small.
On Refact., it also achieves the highest LCU score and improves
FC from 0.7148 to 0.7494 over Meta-Team.
The advantage does not span all dimensions:
ReCreate leads in CQ on both, while Meta-Team leads
in LCU on Feat.
These results highlight consistent strengths in software engineering excellence and functional correctness.

\section{Prompts}
\label{appendix:prompts}

\tcbset{ticerAppendixPrompt/.style={
    ticerAppendixBox,
    enhanced jigsaw,
    unbreakable,
    before upper={\setlength{\parindent}{0pt}\setlength{\parskip}{1pt}\setlength{\emergencystretch}{1em}},
    width=\linewidth,
    arc=2mm,
    boxrule=0.45pt,
    fontupper=\fontsize{10}{11.5}\selectfont,
    fonttitle=\fontsize{10}{11.5}\selectfont\bfseries,
    before skip=0pt, after skip=0pt,
    left=6pt, right=6pt,
    top=4pt, bottom=4pt,
    toptitle=2pt, bottomtitle=2pt
}}

\captionsetup{font=10pt,labelfont=rm,textfont=rm,
    justification=raggedright,singlelinecheck=true,position=bottom}
\begingroup
\makeatletter
\setlength{\@dblfptop}{0pt}
\setlength{\@dblfpbot}{0pt plus 1fil}
\makeatother
\begin{figure*}[!tp]
\centering
\begin{tcolorbox}[ticerAppendixPrompt,
title={Prompt 1: Team Selection},
title after break={Prompt 1: Team Selection (continued)}]
You are a team-selection assistant. Given a task and the team
library, determine whether to reuse an existing agent team or initiate
a cold start.

\textbf{Input:} \textbf{1.} Full task specification; \textbf{2.} Team library.\par

\textbf{Core Rules:}\par\nopagebreak[4]
\begin{list}{\arabic{enumi}.}{%
    \usecounter{enumi}%
    \csname @beginparpenalty\endcsname=10000\relax%
    \setlength{\leftmargin}{1.4em}%
    \setlength{\labelwidth}{1em}%
    \setlength{\labelsep}{0.4em}%
    \setlength{\topsep}{1pt}%
    \setlength{\partopsep}{0pt}%
    \setlength{\itemsep}{1pt}%
    \setlength{\parsep}{0pt}}
    \item Match the task's collaboration pattern, work requirements,
    specialist roles, and verification needs to the available teams.
    \item Choose \textnormal{reuse} only when an existing team supports
    these requirements. Topic or domain similarity alone does not justify reuse.
    \item Choose \textnormal{cold\_start} when no team is suitable. Propose
    a provisional identifier and a short description of the required
    collaboration pattern.
  
    \item Select a team only. Do not recruit members or assign work;
    the global agent pool is not available at this stage.
    \item \ldots
\end{list}
\textbf{Output Format:} Return the decision (\textnormal{reuse} or \textnormal{cold\_start}), team
identifier, and selection rationale. For \textnormal{reuse}, use the existing
identifier. For \textnormal{cold\_start}, also include the provisional description.
\end{tcolorbox}
\nopagebreak[4]
\caption{Prompt for team selection.}
\label{appendix:template_selection_prompt}
\par\vspace{\floatsep}
\centering
\begin{tcolorbox}[ticerAppendixPrompt,
title={Prompt 2: Cold-start Team Composition},
title after break={Prompt 2: Cold-start Team Composition (continued)}]
You are a team-composition assistant. Given a task and a provisional
team description, select candidate members for the cold-start team.

\textbf{Input:} \textbf{1.} Full task specification; \textbf{2.} Provisional team description and selection rationale; \textbf{3.} Fixed Chairman and public agent pool.\par

\textbf{Core Rules:}\par\nopagebreak[4]
\begin{list}{\arabic{enumi}.}{%
    \usecounter{enumi}%
    \csname @beginparpenalty\endcsname=10000\relax%
    \setlength{\leftmargin}{1.4em}%
    \setlength{\labelwidth}{1em}%
    \setlength{\labelsep}{0.4em}%
    \setlength{\topsep}{1pt}%
    \setlength{\partopsep}{0pt}%
    \setlength{\itemsep}{1pt}%
    \setlength{\parsep}{0pt}}
    \item Select a non-empty candidate set that covers the task's
    required capabilities and the provisional collaboration pattern.
    \item Select members only from the supplied agent pool and use their
    existing identifiers. Do not invent agents or identifiers.
    \item Treat the selected members as candidates for recruitment.
    The Chairman decides which candidates to activate for the current task.
    \item Leave work assignments and execution coordination to the Chairman.
    Do not prescribe a workflow or create interaction rules at this stage.
     \item \ldots
\end{list}
\textbf{Output Format:} Return the identifiers of the provisional candidate members.
Do not include the Chairman or task assignments.
\end{tcolorbox}
\nopagebreak[4]
\caption{Prompt for cold-start team composition.}
\label{appendix:cold_start_prompt}
\end{figure*}

\begin{figure*}[!tp]
\centering
\begin{tcolorbox}[ticerAppendixPrompt,
title={Prompt 3: Chairman Coordination},
title after break={Prompt 3: Chairman Coordination (continued)}]
You are the Chairman. Given the task and allowed candidates, recruit and
coordinate a team to complete the task with verifiable results.

\textbf{Input:} \textbf{1.} Full task specification; \textbf{2.} Selected team or provisional candidate set; \textbf{3.} Available teammate profiles; \textbf{4.} Benchmark-specific execution requirements and output format.\par

\textbf{Core Rules:}\par\nopagebreak[4]
\begin{list}{\arabic{enumi}.}{%
    \usecounter{enumi}%
    \csname @beginparpenalty\endcsname=10000\relax%
    \setlength{\leftmargin}{1.4em}%
    \setlength{\labelwidth}{1em}%
    \setlength{\labelsep}{0.4em}%
    \setlength{\topsep}{1pt}%
    \setlength{\partopsep}{0pt}%
    \setlength{\itemsep}{1pt}%
    \setlength{\parsep}{0pt}}
    \item Recruit the smallest sufficient team from the allowed candidates,
    based on the current task's requirements.
    \item Use teammate profiles as supporting evidence. Do not let them
    override the selected team or evidence from the current task.
    \item Assign complementary work with clear objectives, responsibilities,
    and verifiable outcomes.
    \item Coordinate messages and handoffs so that they convey the required
    action, supporting evidence, uncertainty, and verification needs.
    \item Review returned work and reconcile conflicting findings.
    Request further verification or reassign work when progress is blocked
    or the available evidence is insufficient.
    \item Do not infer a lasting capability limitation from one failure
    or from a member not being used.
    \item \ldots
\end{list}
\textbf{Output Format:} During execution, issue recruitment decisions, assignments, messages,
and handoffs through the available interfaces. At completion, submit the
supported task result in the benchmark-required format.
\end{tcolorbox}
\nopagebreak[4]
\caption{Prompt for Chairman coordination.}
\label{appendix:chairman_prompt}
\par\vspace{\floatsep}
\centering
\begin{tcolorbox}[ticerAppendixPrompt,
title={Prompt 4: Interaction-rule Selection and Use},
title after break={Prompt 4: Interaction-rule Selection and Use (continued)}]
You are the sending agent. Given a interaction need, select an applicable
interaction rule and prepare the handoff message.

\textbf{Input:} \textbf{1.} Current interaction need and intended recipient; \textbf{2.} Your active outgoing rules from the selected team's rule library.\par

\textbf{Core Rules:}\par\nopagebreak[4]
\begin{list}{\arabic{enumi}.}{%
    \usecounter{enumi}%
    \csname @beginparpenalty\endcsname=10000\relax%
    \setlength{\leftmargin}{1.4em}%
    \setlength{\labelwidth}{1em}%
    \setlength{\labelsep}{0.4em}%
    \setlength{\topsep}{1pt}%
    \setlength{\partopsep}{0pt}%
    \setlength{\itemsep}{1pt}%
    \setlength{\parsep}{0pt}}
    \item Require the rule's sender and recipient to match the communicating
    agents, its trigger to apply to the current situation, and its instruction
    to match the intended information transfer.
    \item When multiple rules apply, select the most specific one using
    their triggers and evidence requirements to resolve overlaps.
    \item Select one rule per handoff and preserve its existing identifier.
    If no rule applies, select \textnormal{NONE} and communicate directly;
    do not impose an unrelated rule.
    \item Use the selected rule as an information contract, not a fixed
    workflow. Adapt the message to the current task while preserving its
    information, evidence, provenance, uncertainty, verification, and
    fallback requirements.
     \item \ldots
\end{list}
\textbf{Output Format:} Return the intended recipient, the selected rule identifier or
\textnormal{NONE}, and a message stating the required action and supporting
information.
\end{tcolorbox}
\nopagebreak[4]
\caption{Prompt for interaction-rule selection and use.}
\label{appendix:rule_selection_prompt}
\end{figure*}

\begin{figure*}[p]
\centering
\begin{tcolorbox}[ticerAppendixPrompt,
title={Prompt 5: Team Reflection},
title after break={Prompt 5: Team Reflection (continued)}]
You are a team-reflection assistant. Given task execution evidence,
determine whether to create a reusable agent team or revise its candidate
membership, with a corresponding update to its description.

\textbf{Input:} \textbf{1.} Task and current team or cold-start candidate set; \textbf{2.} Actual team and execution summary; \textbf{3.} Evaluation feedback and relevant history.\par

\textbf{Core Rules:}\par\nopagebreak[4]
\begin{list}{\arabic{enumi}.}{%
    \usecounter{enumi}%
    \csname @beginparpenalty\endcsname=10000\relax%
    \setlength{\leftmargin}{1.4em}%
    \setlength{\labelwidth}{1em}%
    \setlength{\labelsep}{0.4em}%
    \setlength{\topsep}{1pt}%
    \setlength{\partopsep}{0pt}%
    \setlength{\itemsep}{1pt}%
    \setlength{\parsep}{0pt}}
    \item Compare the task's requirements with the observed division of work.
    Assess whether the candidate roster supports the required collaboration
    and whether membership changes are warranted.
    \item Describe reusable collaboration patterns, complementary roles,
    evidence and verification principles, and applicability boundaries.
    Do not encode task-specific topics, identifiers, workflows, or tool lists.
    \item For a cold start, choose \textnormal{create} when the experience
    supports a reusable team; choose \textnormal{no\_update} when evidence
    is insufficient.
    \item For an existing team, choose \textnormal{refine} when its
    candidate membership requires revision. Update the candidate roster
    first, then revise the team description to reflect the resulting
    roles, collaboration pattern, and applicability. Choose \textnormal{retain}
    when the current membership remains supported, or \textnormal{no\_update}
    when evidence is insufficient.
    \item Distinguish team-structure problems from execution,
    interaction, and environment failures. Do not revise the team
    merely because a task failed.
    \item Remove a historically useful member only when evidence supports
    persistent mismatch, redundancy, harmful contribution, or a more reliable
    replacement. One failure alone is insufficient.
    \item Treat description changes as accompanying membership updates,
    not as an independent reason to refine a team. Leave agent behavior
    and handoff contracts to their respective reflection modules.
     \item \ldots
\end{list}
\textbf{Output Format:} Return the action and supporting evidence. For a failed task, state whether
team structure contributed to the failure. For \textnormal{create}, provide the
complete candidate-member list and a matching team description. For
\textnormal{refine}, provide the complete revised candidate-member list and
the accompanying revised description. For \textnormal{retain} or
\textnormal{no\_update}, leave the stored membership and description unchanged.
\end{tcolorbox}
\nopagebreak[4]
\caption{Prompt for team reflection.}
\label{appendix:template_reflection_prompt}
\par\vspace{\floatsep}
\centering
\begin{tcolorbox}[ticerAppendixPrompt,
title={Prompt 6: Interaction-rule Reflection},
title after break={Prompt 6: Interaction-rule Reflection (continued)}]
You are an interaction-rule reflection assistant. Given handoff evidence,
assess and update reusable information contracts within the current team.

\textbf{Input:} \textbf{1.} Task, current team, and actual participating agents; \textbf{2.} Handoff traces and rule-use records; \textbf{3.} Current rules, evaluation feedback, and relevant history.\par

\textbf{Core Rules:}\par\nopagebreak[4]
\begin{list}{\arabic{enumi}.}{%
    \usecounter{enumi}%
    \csname @beginparpenalty\endcsname=10000\relax%
    \setlength{\leftmargin}{1.4em}%
    \setlength{\labelwidth}{1em}%
    \setlength{\labelsep}{0.4em}%
    \setlength{\topsep}{1pt}%
    \setlength{\partopsep}{0pt}%
    \setlength{\itemsep}{1pt}%
    \setlength{\parsep}{0pt}}
    \item Scope each rule to a directed sender--recipient pair within
    the current team. Preserve distinct rule identities when the same
    pair requires different contracts under different triggers.
    \item Ground each decision in observed handoffs and outcomes.
    Distinguish information-transfer problems from tool failures,
    missing capabilities, and incorrect task assignments.
    \item Convert task-specific details into reusable requirements for
    information, evidence, provenance, uncertainty, verification, and fallback.
    Express each improvement as actionable handoff guidance.
    \item Choose \textnormal{create} for a useful new contract,
    \textnormal{refine} for a needed revision, or \textnormal{retain} when an
    existing rule remains supported.
    \item Choose \textnormal{retire} only when historical evidence supports
    persistent mismatch, redundancy, harm, ineffective handoffs, or
    replacement by a more reliable rule.
    \item Choose \textnormal{no\_update} when evidence is insufficient.
    Do not turn an isolated failure into an unconditional prohibition.
    \item Update information contracts only. Do not replace team
    composition or agent-skill updates with interaction rules, or turn
    a handoff contract into a fixed task workflow.
     \item \ldots
\end{list}
\textbf{Output Format:} Return the action and supporting evidence. For a rule-specific decision,
provide the rule identifier, sender, recipient, trigger, instruction,
required information and evidence, verification, and fallback. For
\textnormal{refine}, provide a complete revised rule. For \textnormal{no\_update},
leave the rule identifier, sender, and recipient empty.
\end{tcolorbox}
\nopagebreak[4]
\caption{Prompt for interaction-rule reflection.}
\label{appendix:rule_reflection_prompt}
\end{figure*}

\begin{figure*}[!tp]
\centering
\begin{tcolorbox}[ticerAppendixPrompt,
title={Prompt 7: Agent Reflection},
title after break={Prompt 7: Agent Reflection (continued)}]
You are an agent-reflection assistant. Given an agent's execution evidence,
identify reusable improvements to its prompt patches and skills.

\textbf{Input:} \textbf{1.} Task, agent execution trace, and tool calls; \textbf{2.} Final result and evaluation feedback; \textbf{3.} Current prompt patches and skills.\par

\textbf{Core Rules:}\par\nopagebreak[4]
\begin{list}{\arabic{enumi}.}{%
    \usecounter{enumi}%
    \csname @beginparpenalty\endcsname=10000\relax%
    \setlength{\leftmargin}{1.4em}%
    \setlength{\labelwidth}{1em}%
    \setlength{\labelsep}{0.4em}%
    \setlength{\topsep}{1pt}%
    \setlength{\partopsep}{0pt}%
    \setlength{\itemsep}{1pt}%
    \setlength{\parsep}{0pt}}
    \item Ground each proposed improvement in the agent's observed behavior
    and evaluation feedback. Identify whether the issue concerns execution
    behavior, an inadequate prompt, or a missing skill.
    \item Use a prompt patch for a short, behaviorally specific instruction
    that is useful across tasks. Use a skill for a repeatable procedure with
    a clear activation condition, execution steps, and verification.
    \item Do not store one-off answers, task-specific facts or identifiers,
    or temporary workarounds as reusable improvements.
    \item Do not turn team-assignment, handoff, or infrastructure failures
    (model, API, network, tool, Docker, or MCP) into agent updates.
    Leave team membership and handoff contracts to team and rule reflection.
    \item Make the smallest sufficient change. Preserve valid behavior,
    avoid duplicating existing patches or skills, and avoid broad
    instructions that affect unrelated tasks.
    \item Do not generalize from one failure unless the trace reveals
    a repeatable behavioral error. After a success, retain only behavior
    that the trace shows contributed to the result.
    \item Choose one action: \textnormal{create prompt patch},
    \textnormal{refine prompt patch}, \textnormal{create skill},
    \textnormal{refine skill}, \textnormal{retain}, or \textnormal{no update}.
     \item \ldots
\end{list}
\textbf{Output Format:} Return one structured decision. For a create or refine action, include the
trace-grounded reason, observed behavior, reusable lesson, applicability
condition, expected benefit, and limitations. For a prompt patch, provide
its text. For a skill, provide its name, description, when to use, procedure,
verification, and failure handling. Omit task-specific identifiers,
one-off answers, and complete team workflows or handoff protocols.
\end{tcolorbox}
\nopagebreak[4]
\caption{Prompt for agent reflection.}
\label{appendix:agent_reflection_prompt}
\end{figure*}

We present seven core prompts covering team selection, cold-start team
composition, task execution, and post-task reflection. The following
subsections explain each prompt's purpose and role in the overall process.
The corresponding numbered prompts are collected afterward and specify the
role, inputs, core rules, and output format. Repetitive instructions and
tool-call schemas are omitted; benchmark-specific execution requirements
and output formats are appended to each task.
Related engineering guidance emphasizes explicit instructions,
well-defined tool interfaces, and the selection of relevant context for
reliable agent execution and coordination.~\citep{anthropic2024effectiveagents,anthropic2025contextengineering}.

\subsection{Team Selection}
\label{appendix:prompt_description_template_selection}
\noindent
Prompt~1 (Figure~\ref{appendix:template_selection_prompt}) determines whether
an incoming task can reuse an existing agent team or requires a cold start.
The decision is based on the task's specific collaboration requirements, complementary agent roles,
and verification needs, rather than topic similarity alone. This stage
selects the collaboration structure while leaving individual agent recruitment to the Chairman.
When no existing team is suitable, it produces a provisional team description for
subsequent team composition; admission to the team library is decided after task execution.
\subsection{Cold-start Team Composition}
\label{appendix:prompt_description_cold_start}
\noindent
Prompt~2 (Figure~\ref{appendix:cold_start_prompt}) constructs the candidate
roster for a provisional team when team selection returns a cold start.
It selects existing agents whose capabilities support the required division
of work, while keeping the Chairman fixed. The roster defines which agents can
be recruited for the task. Recruitment and work
assignments remain Chairman's responsibility; this prompt does not
prescribe an execution order or generate interaction rules.

\subsection{Chairman Coordination}
\label{appendix:prompt_description_chairman}
\noindent
Prompt~3 (Figure~\ref{appendix:chairman_prompt}) guides task-specific
recruitment and coordination within the selected team's candidate roster.
The Chairman activates the agents needed for the task, assigns complementary
responsibilities
It also reviews returned work, resolves conflicting findings, and adjusts
assignments when necessary. The team defines the available collaboration
structure, while the Chairman instantiates it for the current task and
submits the supported result under the benchmark's output requirements.A related orchestration pattern appears in Anthropic's multi-agent
research system, where a lead agent delegates work to specialized
subagents and synthesizes their findings~\citep{anthropic2025multiagentresearch}.

\subsection{Interaction-rule Selection and Use}
\label{appendix:prompt_description_rule_selection}
\noindent
Prompt~4 (Figure~\ref{appendix:rule_selection_prompt}) guides an agent in
selecting and applying a suitable team-specific interaction rule.
Applicability depends on the communicating agents, the rule's trigger, and
the current information-transfer need. The selected rule specifies what
task information and evidence the message should convey, including uncertainty,
verification, and fallback guidance. If no rule applies, the agent communicates directly without
forcing the handoff to follow an unrelated contract.

\subsection{Team Reflection}
\label{appendix:prompt_description_template_reflection}
\noindent
Prompt~5 (Figure~\ref{appendix:template_reflection_prompt}) evaluates the
team structure after task execution. For a cold start, it assesses whether observed collaboration supports a reusable team that should enter
the library. For an existing team, it determines whether the candidate
membership requires revision and updates the description to match the revised
roster. Description changes accompany membership updates. Decisions are
grounded in execution evidence and relevant history, distinguishing structural
mismatches from execution, interaction, and environment failures.

\subsection{Interaction-rule Reflection}
\label{appendix:prompt_description_rule_reflection}
\begingroup
\widowpenalty=10000
\brokenpenalty=10000
\noindent
Prompt~6 (Figure~\ref{appendix:rule_reflection_prompt}) converts observed
handoffs and evaluation feedback into reusable interaction-rule updates
within the current team. It assesses directed sender--recipient contracts
and decides whether to create, refine, retain, or retire a rule, or leave
the rule library unchanged when evidence is insufficient. The reflection
focuses on information-transfer requirements and preserves distinct triggers
for rules serving the same agent pair. Team membership and individual
agent behavior are handled by their respective reflection prompts.
\par\endgroup

\subsection{Agent Reflection}
\label{appendix:prompt_description_agent_reflection}
\noindent
Prompt~7 (Figure~\ref{appendix:agent_reflection_prompt}) extracts reusable
improvements to an individual agent's execution behavior from its trace and
evaluation feedback. Short behavioral instructions are stored as prompt
patches, while repeatable procedures with explicit activation conditions and
verification steps are stored as skills. The prompt favors small, supported
updates that preserve existing valid behavior and avoid duplication. It excludes
one-off task answers and does not attribute team-assignment, handoff, or
infrastructure failures to the agent's reusable capabilities.Reusable procedural knowledge is also central to Agent Skills, which
packages instructions, scripts, and supporting resources for agents to
discover and load when needed~\citep{anthropic2025agentskills}.
\begin{figure*}[!t]
    \centering
    \includegraphics[width=0.99\textwidth]
    {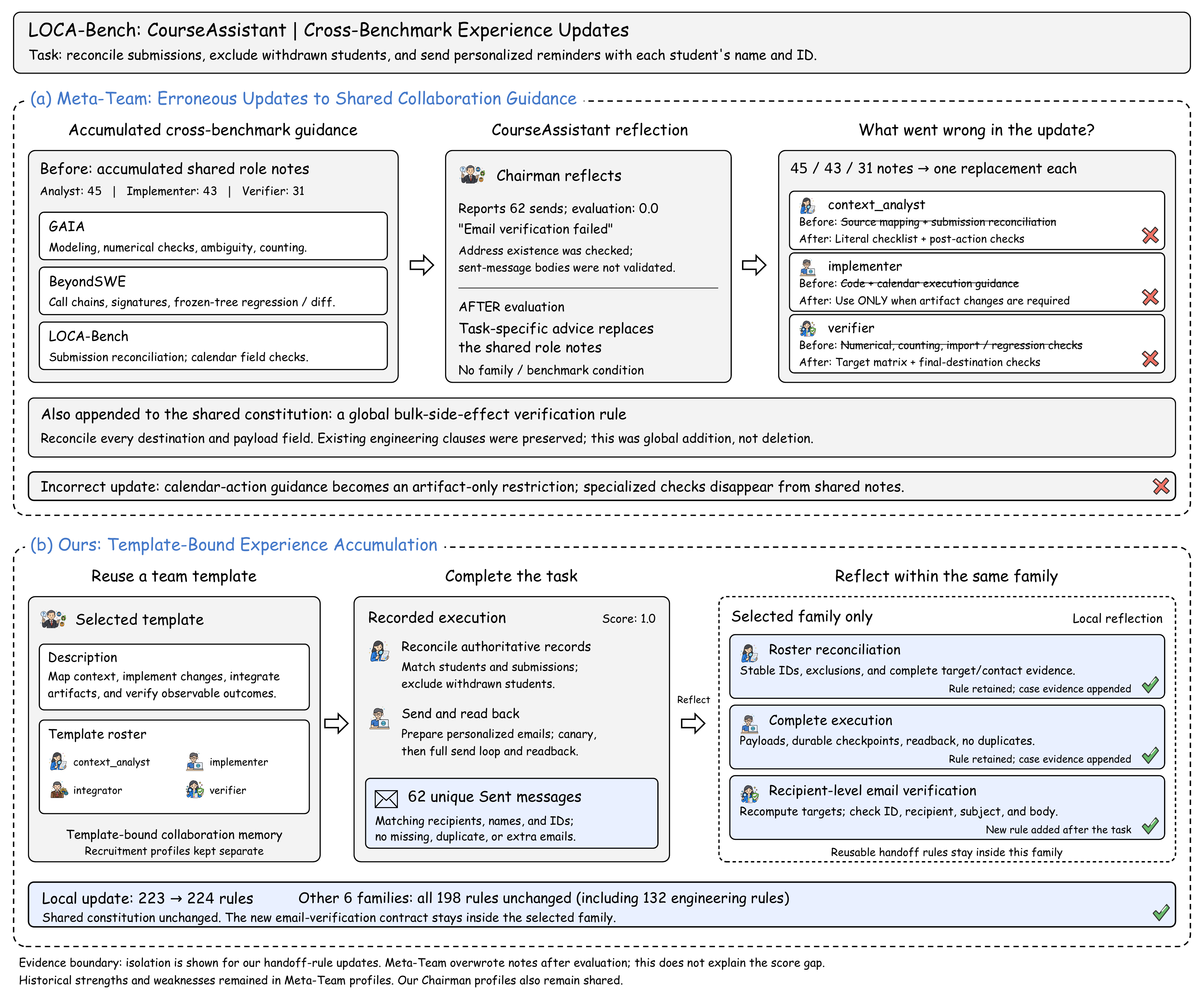}
    \caption{
    Family-scoped experience accumulation versus shared-profile
    contamination on CourseAssistant.
    (a) Meta-Team replaces accumulated shared-profile notes with
    task-derived advice, removing earlier guidance and introducing
    an unscoped recruitment restriction.
    Strikethroughs indicate overwritten guidance; crosses mark
    problematic replacements.
    (b) \textsc{EvoTeam} preserves existing rules, appends supporting
    evidence, and adds a verification rule within the selected family.
    All 198 rules in the other six families remain unchanged.
    }
    \label{fig:courseassistant-scoped-updates}
\end{figure*}

\begin{figure*}[t]
    \centering
    \includegraphics[width=\textwidth]
    {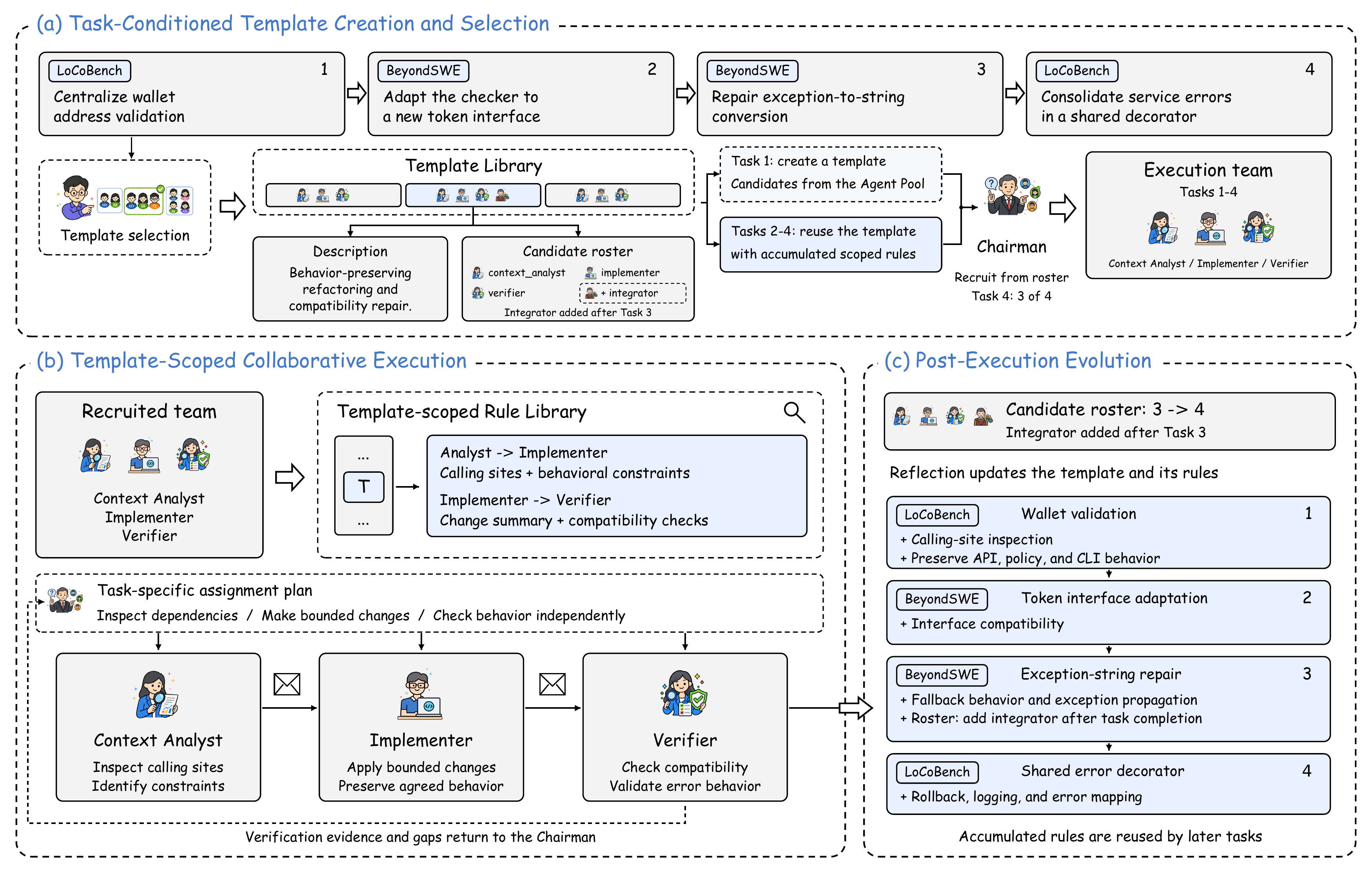}
    \caption{Cross-benchmark collaboration evolution.
    (a) Team creation and reuse.
    (b) Collaboration guided by team-scoped interaction rules.
    (c) Accumulated rule updates and candidate-roster expansion
    across four tasks.}
    \label{fig:cross_benchmark_evolution}
\end{figure*}

\begin{figure*}[t]
\centering
\begin{minipage}{\textwidth}

\fontsize{10}{12}\selectfont
\centering
\setlength{\tabcolsep}{5pt}
\renewcommand{\arraystretch}{1.2}

\captionof{table}{Simplified trace of the Case~3 escrow development task (LoCoBench).}
\label{tab:case4_trace}

\begin{tabularx}{\textwidth}{
    >{\raggedright\arraybackslash}p{0.13\textwidth}
    >{\raggedright\arraybackslash}p{0.17\textwidth}
    >{\raggedright\arraybackslash}X}
\toprule
\textbf{Agent} & \textbf{Action} & \textbf{Key content} \\
\midrule
Context Analyst & Inventory structure & Maps interfaces across the domain, workflow, API, and risk layers; identifies state transitions and integration constraints. \\
Implementer & Cross-layer construction & Implements the escrow feature across the affected layers while preserving state contracts and existing interfaces. \\
Verifier & Independent verification & Independently checks compilation, workflow behavior, boundary conditions, and cross-layer consistency. \\
\bottomrule
\end{tabularx}

\end{minipage}
\par\vspace{\floatsep}
\begin{minipage}{\textwidth}
\begin{tcolorbox}[ticerAppendixPrompt,
    width=\linewidth,
    title={Scoped rule: independent verification of cross-layer changes},
    before skip=0pt, after skip=0pt]
\textbf{Scope.} Features that span multiple components and require consistent
interfaces, state transitions, and access control.

\textbf{Handoff requirement.} The verifier receives the interface inventory
and a summary of the implemented changes, including affected components and
expected behavior.

\textbf{Verification responsibility.} Independently check the implementation
against these contracts. Review cross-component consistency and exercise the
relevant state transitions, authorization boundaries, and failure cases.

\textbf{Required evidence.} Report the components reviewed, tests performed,
results obtained, and unresolved gaps. Distinguish compilation and static
checks from runtime evidence.

\textbf{Unavailable dependencies.} Identify blocked checks and their causes;
do not infer runtime correctness from compilation or imports alone.
\end{tcolorbox}
\captionsetup{font=10pt,position=bottom}
\captionof{figure}{Condensed scoped rule for independent verification of cross-layer
changes. Explicit handoff requirements connect implementation constraints to
verifiable evidence.}
\label{fig:case4_rule}
\end{minipage}
\par\vspace{\floatsep}
\begin{minipage}{\textwidth}
\begin{tcolorbox}[ticerAppendixPrompt,
    width=\linewidth,
    title={Agent team: repository-grounded build and validation},
    before skip=0pt, after skip=0pt]
\textbf{Description.}
A repository-grounded build-and-validation collaboration in which one member
maps structure and dependencies, another produces the change, an integrator
reconciles cross-component contracts and keeps the result coherent, and a
verifier independently checks behavior against explicit requirements.
Evidence comes from inspected artifacts, focused tests, and end-to-end
consistency checks rather than assumptions; this pattern fits multi-surface
software changes and other deliverables where integration risk matters, but
is unnecessary for isolated edits or purely exploratory analysis.

\textbf{Candidate members.}
Context Analyst, Implementer, Verifier, and Integrator.
\end{tcolorbox}
\captionsetup{font=10pt,position=bottom}
\captionof{figure}{The shared agent team reused across the three tasks in Case~3.
The roster specifies candidate members for task-specific recruitment.}
\label{fig:case4_template}
\end{minipage}
\end{figure*}

\raggedbottom
\clubpenalty=10000
\widowpenalty=10000
\setlength{\emergencystretch}{1em}
\section{Case Study}
\label{appendix:case_study}

We present four cases drawn from the mixed-benchmark evolution and holdout
splits. Cases 1 and 2 concern what is retained during
evolution: Case 1 shows that a task-derived update stays confined to the
selected team's rule set instead of overwriting experience accumulated for
other collaboration patterns, and Case 2 follows one team across an
interleaved LoCoBench and BeyondSWE sequence. Cases 3 and 4 use the frozen
library on the holdout split: Case 3 shows three tasks from three different benchmarks
selecting the same team, and Case 4 compares \textsc{EvoTeam} with Meta-Team on
a single task, tracing the score difference to specific team configuration and handoff
requirements.

\subsection{Scoped Experience Accumulation}
\label{appendix:case 4}

\paragraph{Task and comparison.}
Figure~\ref{fig:courseassistant-scoped-updates} compares
post-execution updates on CourseAssistant from LOCA-Bench.
The task requires reconciling student and submission records,
excluding withdrawn students, and sending individualized reminders
with a prescribed subject and each student's name and ID.
The comparison focuses on how task-derived experience affects
existing collaboration knowledge previously accumulated across benchmarks.

\paragraph{Meta-Team: overwriting shared guidance.}
Before this task, the shared profiles of
context analyst, implementer, and
verifier contained 45, 43, and 31 notes, respectively.
These covered numerical reasoning from GAIA, repository analysis
and code verification from BeyondSWE, and record reconciliation
and calendar operations from LOCA-Bench.
The Chairman reported 62 successful sends, but evaluation returned
0.0 with ``Email verification failed.''
Reflection identified that address existence had been checked
without validating sent-message bodies.
It then replaced each accumulated notes list with a single note.
Specialized analysis and verification requirements disappeared,
while the implementer's new advice restricted recruitment to
tasks requiring artifact changes, despite earlier guidance
covering calendar operations.
These replacements were shared by teammate name without a
benchmark or family condition.
A bulk-side-effect verification rule was appended to the
shared constitution, retaining its engineering clauses.

\paragraph{\textsc{EvoTeam}: retaining and extending scoped experience.}
The selected team supports structural analysis, implementation,
artifact reconciliation, and independent outcome verification,
with a candidate roster of context\_analyst,
implementer, integrator, and verifier.
Execution reconciled authoritative records and performed
individualized sending with readback.
The final record reported 62 unique Sent messages with matching
recipients, names, and IDs, without missing, duplicate, or extra
messages; evaluation returned 1.0.
Subsequent reflection retained the roster-reconciliation and
complete-execution rules and appended this case as supporting evidence.
It added one family-scoped rule requiring independent target-set
reconstruction and recipient-level checks of message IDs,
recipients, subjects, and required body fields.

\paragraph{Observed distinction.}
Meta-Team's profile-note replacement removed accumulated
distinctions and generalized task-derived advice across future tasks.
\textsc{EvoTeam}'s handoff-rule update incorporated new evidence while
preserving rules outside the selected family.
This isolation concerns handoff rules: our recruitment profiles
remain shared, and Meta-Team retained historical strengths and
weaknesses outside the overwritten notes.
Because the reflection process followed evaluation, the overwrite does not
establish the cause of the task-score difference.

\subsection{Cross-Benchmark Accumulation}
\label{appendix:cross_benchmark_case}


Figure~\ref{fig:cross_benchmark_evolution} illustrates how
a shared team accumulates reusable collaboration experience
across four distinct tasks from LoCoBench and BeyondSWE in the
mixed-benchmark evolution stream.
The team preserves shared coordination practices while adapting
agent recruitment and rule application to the collaboration
requirements of each task.

\paragraph{Team creation and reuse.}
In panel (a), the wallet-validation task initiates a
provisional team for behavior-preserving refactoring
and compatibility repair, which is retained and updated
through post-execution reflection.
Subsequent tasks reuse this team based on their
collaboration requirements.
Its description defines the applicability scope, while
its candidate roster specifies the agents available
for recruitment by the Chairman.

\paragraph{Team-scoped collaboration.}
Panel (b) illustrates the shared division of work.
The context analyst inspects calling sites and identifies
behavioral constraints; the implementer makes bounded
changes that preserve these constraints; and the verifier
independently checks compatibility and error behavior.
Team-scoped rules specify the information required at each handoff,
including calling sites, behavioral constraints, change summaries,
and compatibility checks.
Verification evidence and unresolved gaps are returned
to the Chairman.
These rules guide information flow under task-specific plans 
rather than impose a fixed execution sequence on the team.

\paragraph{Experience accumulation.}
Panel (c) summarizes the updates contributed by the
four tasks: calling-site inspection and behavior
preservation; interface compatibility; fallback behavior
and exception propagation; and rollback, logging,
and error mapping.
After the third task, reflection adds an integrator
to the candidate roster.
The fourth task reuses the expanded team but recruits
only the context analyst, implementer, and verifier.
This illustrates how a team retains collaboration
experience across benchmarks while providing a candidate
roster from which the Chairman selects the members
needed to meet each task’s specific requirements.

\subsection{Cross-Benchmark Team Reuse}
\label{appendix:cross_benchmark_template_reuse}

\begin{figure*}[t]
    \centering
    \includegraphics[width=\textwidth]
    {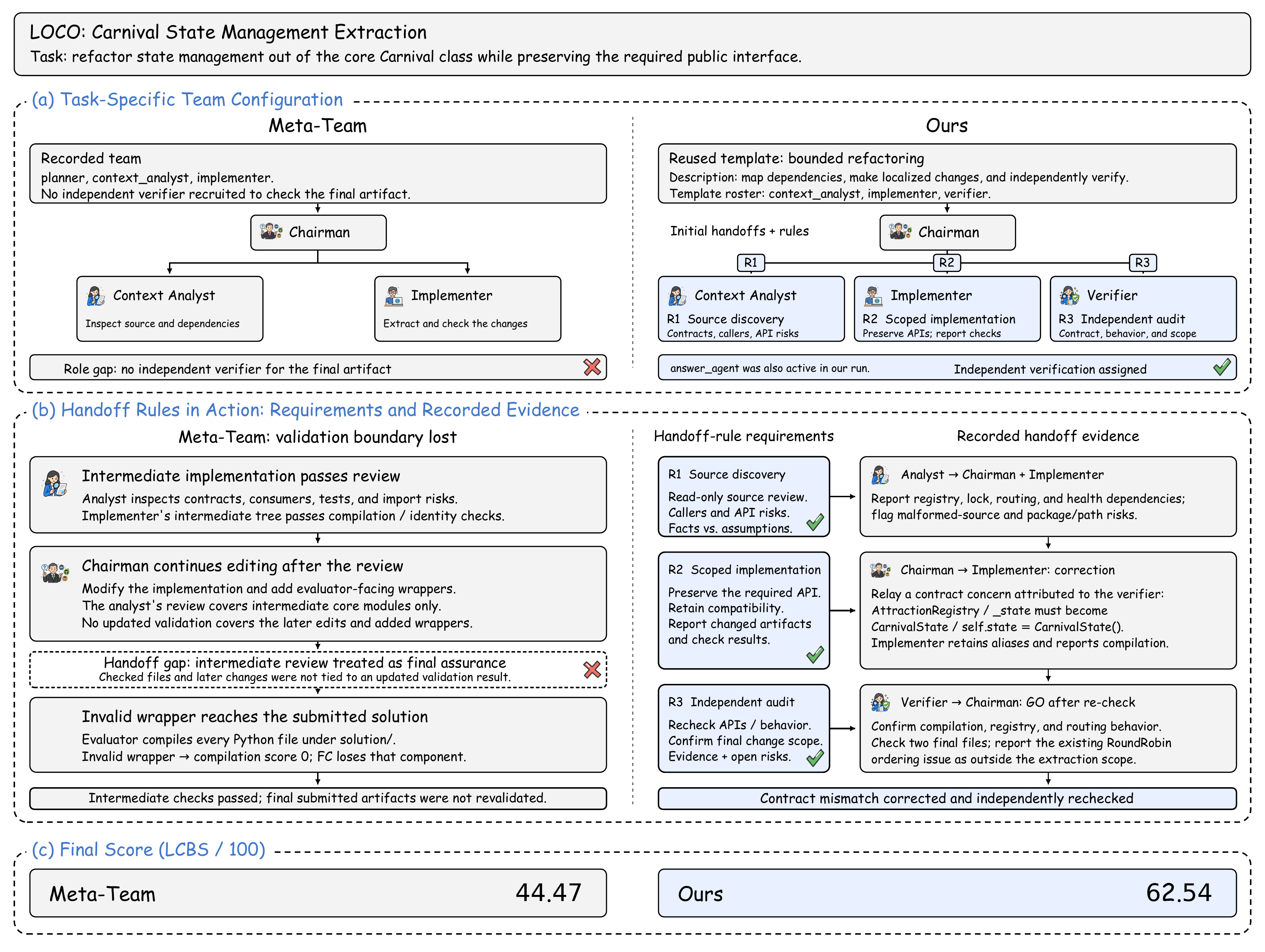}
    \caption{\textbf{Team configuration and rule-guided
    collaboration on the Carnival refactoring task.}
    (a) Our method recruits an independent verifier and
    attaches team-scoped rules to the initial handoffs.
    Meta-Team recruits the analyst and implementer without
    a separate verifier.
    (b) Our team corrects an interface mismatch and
    independently rechecks the revised files.
    Meta-Team's later edits lack updated validation,
    leaving an invalid wrapper in the submission.
    (c) Final LCBS scores (out of 100).}
    \label{fig:carnival_comparison}
\end{figure*}

Three holdout tasks reuse the same team for dependency analysis, coordinated implementation, and independent verification. 
 They involve escrow feature implementation (LoCoBench), a clearance
promotion workflow (LOCA-Bench), and ORM compatibility repair (BeyondSWE). These tasks activate only
the context analyst, implementer, and verifier. Their scoped rules make key
cross-component constraints explicit at each team handoff.

\begin{ticerItemize}
\item \textbf{Escrow development (LoCoBench).} The analyst identifies the
interfaces and state transitions needed for a time-locked multi-signature
escrow feature. The implementer develops the feature across the domain,
workflow, API, and risk layers while preserving those contracts. The verifier
independently checks compilation, workflow behavior, and boundary conditions.
The handoff rules require an interface inventory before implementation and clear
verification evidence for the affected components afterward.

\item \textbf{Clearance promotion (LOCA-Bench).} The analyst checks product,
order, and subscriber data and establishes the required CSV format. The
implementer updates the clearance category, generates the CSV, and emails
subscribers. The verifier checks product selection, file contents, recipients,
and CSV delivery. These contract-preservation 
and independent verification rules govern a product–file–email workflow.

\item \textbf{Compatibility repair (BeyondSWE).} The analyst identifies how
SQLAlchemy changes affect SQLAdmin's relationship loading. The implementer
limits edits to the affected compatibility layer, and the verifier reruns
regression tests and distinguishes dependency problems from code regressions.
The scoped rules connect an explicit impact analysis to bounded changes and
independent validation of those changes.
\end{ticerItemize}

All three tasks succeed. The shared team transfers
because these tasks share a coordination pattern: analysis establishes
constraints, implementation preserves agreed contracts, and verification
checks the result independently. Table~\ref{tab:case4_trace} summarizes the
escrow task, while Figure~\ref{fig:case4_rule} illustrates
the team's verification rule during execution.

\subsection{Organization and Coordination}
\label{appendix:carnival_case}

This case illustrates how a task-appropriate team and team-scoped
interaction rules support collaboration on the Carnival refactoring task
(Figure~\ref{fig:carnival_comparison}), which requires structural changes
while ensuring that the existing interfaces are preserved.

\paragraph{Matching the team to the task.}
\textsc{EvoTeam} reuses a team that combines dependency analysis,
bounded implementation, and independent verification.
The Chairman recruits a context analyst, implementer,
and verifier from its own roster, covering the task’s requirements
with a clear division of responsibilities.
Meta-Team recruits an analyst and an implementer but no independent
verifier, leaving final validation without a dedicated role.

\paragraph{Connecting responsibilities through rules.}
The team-scoped rules make this division of work operational:
compatibility constraints must guide implementation, changed artifacts
and check results must be reported, and independent verification
must cover the revised deliverables.
In the recorded execution, an interface mismatch is corrected and
the revised files receive an independent audit.
Meta-Team performs useful intermediate checks, but subsequent
modifications are delivered without updated validation, leaving
an invalid wrapper in the submission.
The contrast highlights the complementary roles of assigning
verification responsibility and specifying what must be verified.

\textsc{EvoTeam} achieves an LCBS of 62.54, compared with 44.47
for Meta-Team (panel~(c)).
This case illustrates how a team matched to the task provides
the needed roles, while scoped interaction rules connect their respective contributions
to support a more reliable final result for the task.
\section{Extended Related Work}
\label{appendix:related_work}

\subsection{Individual Agent Evolution}

Research on agent improvement explores how feedback can
refine task-solving behavior and produce reusable knowledge~\cite{xiang2026systematic}.
Early approaches establish feedback-based refinement
through iterative output revision and reflection-guided
repeated attempts, with different scopes of experience
retention~\citep{madaan2023selfrefineiterativerefinementselffeedback,shinn2023reflexionlanguageagentsverbal}.
Self-Refine improves generated outputs through
self-generated feedback, while Reflexion retains verbal
reflections in episodic memory to guide subsequent trials.
ExpeL extends experience reuse across tasks by extracting
natural-language insights from past executions and
recalling these insights alongside the relevant previous experiences during subsequent task
solving~\citep{zhao2024expelllmagentsexperiential}.

Beyond response-level refinement, automated agent
optimization extends to code-defined designs, functional
modules, and symbolic components such as prompts and
tools~\citep{hu2025automateddesignagenticsystems,shang2025agentsquareautomaticllmagent,zhou2024symboliclearningenablesselfevolving}.
ADAS uses a meta-agent to propose new agent designs based on an
archive of previous discoveries, while AgentSquare evolves
and recombines reusable planning, reasoning, tool-use, and memory
modules.
Agent Symbolic Learning further optimizes prompts, tools,
and their overall composition through natural-language feedback
and symbolic updates to these components.

Another line of work converts interaction experience into
reusable procedural knowledge and revised execution
scaffolds~\citep{wang2023voyageropenendedembodiedagent,zheng2025skillweaverwebagentsselfimprove,hao2026recreatereasoningcreatingdomain}.
Voyager accumulates executable skills through open-ended
environment interaction, while SkillWeaver enables web
agents to discover and practice skills, distilling
interaction experience into reusable APIs.
ReCreate reasons over interaction histories to identify
causes of success and failure, translating execution
experience into scaffold revisions and reusable domain
patterns.
Mem$^2$Evolve connects experience accumulation with
capability expansion by using experience memory to guide
the creation of reusable tools and expert
agents~\citep{cheng-etal-2026-mem2evolve}.
Agentic Context Engineering (ACE) instead maintains evolving context
playbooks through generation, reflection, and curation, using incremental
updates to preserve useful knowledge across
executions~\citep{zhang2026agenticcontextengineering}.
Within multi-agent collaboration, MAEL maintains individual
experience pools and retrieves high-reward, task-relevant
examples to improve agents' subsequent reasoning
steps~\citep{li2025crosstaskexperientiallearningllmbased}.
Together, these studies improve agent designs and retain
reflections, skills, and execution knowledge beyond a
single response.
However, analyses of multi-agent failures and scaling
behavior show that collaborative performance also depends
on inter-agent alignment and the compatibility between
coordination structures and task
requirements~\citep{cemri2025multiagentllmsystemsfail,kim2026sciencescalingagentsystems}.
This leaves a distinct question for collaborative
evolution: how should accumulated capabilities be
organized and coordinated when tasks require different
responsibilities and information handoffs?

\subsection{Collaborative Multi-Agent Evolution}

At the collaboration level, improvement extends to
interactions and the structures connecting agent behaviors.
Studies of large-scale collaboration highlight the role
of interaction topology in shaping collective
performance~\citep{qian2025scalinglargelanguagemodelbased,kim2026sciencescalingagentsystems}.
Related approaches adjust team composition and agent
participation to improve task solving, including dynamic
group composition, contribution-based agent selection,
and elimination of redundant agents and
interaction~\citep{chen2023agentversefacilitatingmultiagentcollaboration,liu2024dynamicllmpoweredagentnetwork,wang-etal-2025-agentdropout,liang2022mas}.
These approaches provide mechanisms for adapting
collaboration to a task's requirements.

Automated optimization further treats agent prompts,
graph connectivity, and executable workflows as
searchable components~\citep{pmlr-v235-zhuge24a,zhou2026multiagentdesignoptimizingagents,zhang2025aflowautomatingagenticworkflow}.
GPTSwarm optimizes node-level prompts and graph
connectivity, MASS interleaves prompt and topology
optimization, and AFlow searches over code-represented
workflows using execution feedback.EvoMAC uses textual backpropagation from environmental feedback to
revise multi-agent collaboration networks, while EvoFlow evolves a
population of diverse workflows through retrieval, crossover, and
mutation~\citep{hu2025evomac,zhang2025evoflow}.
ScoreFlow extends workflow optimization through
score-based preference learning, while EvoAgent uses
evolutionary operators to generate diverse agents from
an initial agent framework~\citep{wang2025scoreflowmasteringllmagent,yuan2025evoagentautomaticmultiagentgeneration}.
To accommodate varying task requirements, query-dependent
approaches learn to select or generate collaboration
architectures, agent roles, and model assignments
according to the input
query~\citep{pmlr-v267-zhang25bi,gao2025flowreasonerreinforcingquerylevelmetaagents,yue2025masrouter}.
These methods make collaboration structures explicit
optimization targets and support task-dependent
system construction.

Experience-driven approaches investigate how
execution knowledge can support subsequent collaboration,
including retrieval-based agent memory, collective
reflection, and reusable orchestration
knowledge~\citep{zhang2026skillpoison,zhou2026eviskill,yang2026graph,ning2026memadapter}.
AgentNet combines retrieval-based memory with decentralized
coordination, adapting connectivity and task routing
according to local expertise and context.
DecentMem gives each agent separate memory pools for consolidated
experience and generated exploratory candidates, adjusting their
relative weights through online feedback~\citep{hao2026decentmem}.
Meta-Team preserves individual execution contexts and
enables post-task evidence exchange, supporting evolution
of agent behaviors, inter-agent coordination, and team
organization.
Skill-based approaches connect reusable knowledge with
changes to multi-agent collaboration: SkillMAS couples
skill evolution with evidence-guided MAS restructuring,
while Skill-MAS represents orchestration knowledge as an
evolvable Meta-Skill refined through multi-trajectory
rollout and selective
reflection~\citep{lin2026skillmasevolvingmetaskillautomatic}.
EMAS uses recurring execution diagnoses to propose prompt
and topology revisions, accepting updates through paired
validation against the current
system~\citep{fei2026emasstabilizingmultiagentevolution}.
MASFly and MANTA combine prior experience with
execution feedback to support task-specific collaboration
structures and execution-time
reconfiguration.

These studies demonstrate several ways to improve
collaboration through structural optimization and
experience reuse~\citep{feng2026graph}.
Our work focuses on sustaining such improvement across
heterogeneous task requirements, where coordination
knowledge must remain appropriate to the team using it.
\textsc{EvoTeam} maintains a library of reusable teams,
selecting or creating teams according to collaboration
requirements.
It jointly refines team organization and associated
interaction rules, allowing compatible tasks to accumulate
coordination experience within the same team while keeping
rule updates independent across teams.
Individual agent skills remain reusable across these
team contexts.

\section{Frequently Asked Questions}
\label{appendix:qa}
\paragraph{Q1. Is EvoTeam only a task-dependent team-selection method?}
\textbf{A.}
Team selection is the entry point to a broader process of
collaboration evolution.
A reusable team specifies supported collaboration requirements
and a candidate roster, from which the Chairman recruits a
task-specific execution team.
Tasks can therefore reuse the same team while activating
different members and assigning different responsibilities.
Beyond selection, \textsc{EvoTeam} uses execution experience
to refine team organization, associated interaction rules,
and individual agent skills.
The reusable team provides a persistent context in which
tasks with compatible collaboration requirements accumulate
and refine coordination experience
(Section~\ref{sec:method};
Algorithm~\ref{alg:team_evolution}).

\paragraph{Q2. Are teams organized by benchmark or task category?}
\textbf{A.}
Teams are organized according to collaboration requirements,
rather than predefined benchmark or task-category boundaries.
The LLM-based Team Selector assesses whether a team's
description and candidate roster support the required division
of work, complementary roles, and verification needs.
Topic similarity alone is insufficient to justify reuse.
Consequently, tasks from different benchmarks can reuse the
same team when their collaboration requirements are compatible,
as illustrated in Appendix~\ref{appendix:cross_benchmark_template_reuse}.
When no compatible team exists during evolution, the Composer
constructs a provisional team from the global agent pool.
Post-execution reflection determines its admission based on
whether the roster supports the required collaboration and
the resulting organization is reusable
(Section~\ref{sec:method_template}).

\paragraph{Q3. Do interaction rules impose a fixed execution workflow?}
\textbf{A.}
Interaction rules provide soft coordination guidance about
the information and evidence required at a handoff.
They do not prescribe a complete execution sequence.
The Chairman establishes task-specific assignments and can
reassign subtasks when progress is blocked.
For each handoff, the sending agent assesses rule applicability
using the sender--recipient pair, the rule's trigger, and
the current information-transfer requirements.
When a rule applies, the agent adapts its message to the
current task while following the relevant information and
verification requirements.
If no rule applies, the agent communicates directly
without imposing an unrelated rule
(Section~\ref{sec:task-execution};
Appendix~\ref{appendix:prompt_description_rule_selection}).

\paragraph{Q4. Why are interaction rules team-specific while
agent skills remain reusable across teams?}
\textbf{A.}
Interaction rules depend on how responsibilities and
information dependencies are organized within a team.
A handoff requirement that supports one collaboration
pattern may be inappropriate for another.
Individual skills instead encode procedures for an agent's
own execution and can remain useful across teams when their
applicability conditions are satisfied.
\textsc{EvoTeam} therefore updates coordination knowledge
within the corresponding team while retaining skills with
the agents that use them.
This separation allows compatible tasks to share and refine
interaction rules without directly modifying other teams'
rule sets.
The isolation applies to team-specific rules; individual
skills and recruitment profiles remain shared across team
contexts
(Section~\ref{sec:method_evolution};
Appendix~\ref{appendix:case 4}).

\paragraph{Q5. What evolves, and does evolution continue during
holdout evaluation?}
\textbf{A.}
Evolution updates three forms of persistent knowledge:
team candidate rosters and their corresponding descriptions,
team-specific interaction rules, and individual agent guidance.
The latter includes prompt patches and reusable skills in
our implementation.
The underlying LLM parameters remain fixed.
Persistent updates are performed on the designated evolution
split and frozen before evaluation on the disjoint holdout split.
Task-specific recruitment, planning, rule application, and
reassignment remain part of execution.
Thus, freezing evolution prevents holdout tasks from changing
the persistent knowledge used by subsequent evaluation tasks,
while preserving adaptation within each task
(Algorithm~\ref{alg:team_evolution};
Appendices~\ref{appendix:execution_evolution_procedure}
and~\ref{appendix:prompt_description_agent_reflection}).

\paragraph{Q6. Does the team library support cross-benchmark reuse, or merely isolate experience for individual tasks?}
\textbf{A.} Teams are organized by collaboration requirements rather than task identities or benchmark labels. Appendix~\ref{appendix:cross_benchmark_case} follows one team across four LoCoBench and BeyondSWE tasks, including accumulated rule updates and candidate-roster expansion. Appendix~\ref{appendix:cross_benchmark_template_reuse} further shows a shared team reused for holdout tasks from three benchmarks. Together, these cases illustrate how compatible tasks across benchmarks accumulate and reuse coordination experience within a shared collaboration structure.

\paragraph{Q7. How do the ablation studies and case analyses support team-specific interaction rules?}
\textbf{A.} The ablation studies show that learned interaction rules contribute to performance beyond task-conditioned team organization. Appendix~\ref{appendix:case 4} further illustrates how task-derived coordination experience updates the selected team's rules while leaving other teams' rules unchanged. These analyses provide complementary evidence: the ablations assess the value of learned coordination guidance, while the case analysis illustrates how team-scoped evolution incorporates new experience while preserving interaction rules associated with other collaboration structures.

\paragraph{Q8. What do the reported comparisons show about EvoTeam's performance--cost trade-off?}
\textbf{A.} Figure~\ref{fig:mixed_cost_performance} and Figure~\ref{fig:cost} report task-level performance--cost comparisons in the evaluated settings. Under mixed-benchmark evolution, \textsc{EvoTeam} achieves stronger performance at lower reported per-task cost than Meta-Team and ReCreate on GAIA and BeyondSWE DepMig. These results show that \textsc{EvoTeam} combines strong task performance with competitive per-task costs, supporting its practical effectiveness across different collaboration requirements.

\end{document}